\documentclass[10pt,twocolumn,letterpaper]{article}

\usepackage[pagenumbers]{cvpr} % To force page numbers, e.g. for an arXiv version

\usepackage{lineno}
\definecolor{color_generated_frame}{RGB}{0,204,204}
\usepackage{graphicx} % Add this to your preamble if not already included

\definecolor{cvprblue}{rgb}{0.21,0.49,0.74}
\usepackage[pagebackref,breaklinks,colorlinks,allcolors=cvprblue]{hyperref}

\def\paperID{33} % *** Enter the Paper ID here
\def\confName{ICCV}
\def\confYear{2025}

\usepackage{amsmath,amsfonts,bm}

\def\eqref#1{equation~\ref{#1}}
\def\1{\bm{1}}

\DeclareMathAlphabet{\mathsfit}{\encodingdefault}{\sfdefault}{m}{sl}
\SetMathAlphabet{\mathsfit}{bold}{\encodingdefault}{\sfdefault}{bx}{n}

\newcommand{\E}{\mathbb{E}}

\usepackage{graphicx}
\usepackage{amsmath}
\usepackage{amssymb}
\usepackage{booktabs}
\usepackage{times}
\usepackage{microtype}
\usepackage{epsfig}
\usepackage{caption}
\usepackage{xcolor}
\usepackage{float}
\usepackage{placeins}
\usepackage{color, colortbl}
\usepackage{stfloats}
\usepackage{enumitem}
\usepackage{tabularx}
\usepackage{xstring}
\usepackage{multirow}
\usepackage{xspace}
\usepackage{url}
\usepackage{subcaption}
\usepackage{arydshln}
\usepackage{xcolor}
\usepackage{bbm}
\usepackage[hang,flushmargin]{footmisc}
\usepackage{pifont}%
\newcommand{\cmark}{\ding{51}}%
\newcommand{\xmark}{\ding{55}}%
\usepackage{hyperref}
\usepackage{wrapfig}
\usepackage{lipsum}
\usepackage{comment}
\usepackage{algorithm}
\usepackage{algorithmic}
\usepackage{listings}
\usepackage{makecell} 
\usepackage{xcolor}
\definecolor{lightblue}{RGB}{185, 200, 240}
\usepackage{pifont}
\usepackage{tikz}
\usepackage{makecell}

\usepackage[capitalize]{cleveref}
\crefname{section}{Sec.}{Secs.}
\Crefname{section}{Section}{Sections}
\Crefname{table}{Table}{Tables}
\crefname{table}{Tab.}{Tabs.}

\usepackage{wrapfig}

\newcommand{\denoise}{p_{1|t}}

\title{
MaskFlow: Discrete Flows For Flexible and Efficient Long Video Generation
}

\author{Michael Fuest, Vincent Tao Hu\thanks{ Project Leader}, Björn Ommer \\
{ CompVis @ LMU Munich, MCML}\\
\\
\url{https://compvis.github.io/maskflow/}
}
\vspace{-10pt}

\begin{document}
\maketitle

\begin{abstract}
     Generating long, high-quality videos remains a challenge due to the complex interplay of spatial and temporal dynamics and hardware limitations. In this work, we introduce \textit{MaskFlow}, a unified video generation framework that combines discrete representations with flow-matching to enable efficient generation of high-quality long videos. By leveraging a frame-level masking strategy during training, MaskFlow conditions on previously generated unmasked frames to generate videos with lengths ten times beyond that of the training sequences. MaskFlow does so very efficiently by enabling the use of fast Masked Generative Model (MGM)-style sampling and can be deployed in both fully autoregressive as well as full-sequence generation modes. We validate the quality of our method on the FaceForensics (FFS) and Deepmind Lab (DMLab) datasets and report Fréchet Video Distance (FVD) competitive with state-of-the-art approaches. We also provide a detailed analysis on the sampling efficiency of our method and demonstrate that MaskFlow can be applied to both timestep-dependent and timestep-independent models in a training-free manner.

\end{abstract}

\begin{figure}
    \centering
    \begin{tikzpicture}[font=\footnotesize]
        \node (img) {\includegraphics[width=0.7\columnwidth]{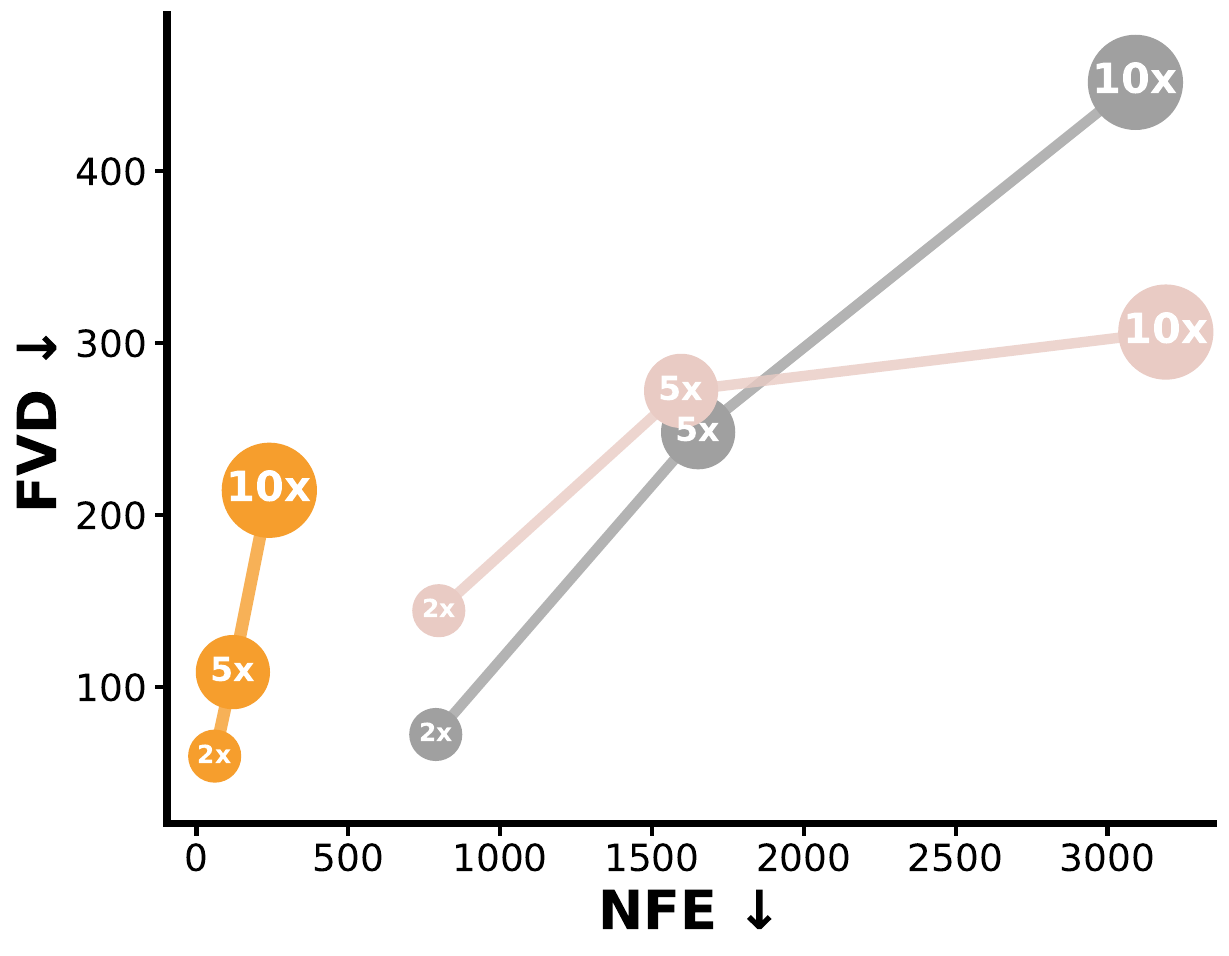}};
            \node[anchor=north west, xshift=25pt, yshift=-5pt] at (img.north west) {
                \begin{tabular}{ll}
                \scriptsize
                    \textcolor[HTML]{A0A0A0}{\rule{6pt}{6pt}} &Rolling Diffusion \cite{ruhe2024rollingdiffusionmodels} \\
                    \textcolor[HTML]{e9cbc4}{\rule{6pt}{6pt}} &Diffusion Forcing \cite{chen2024diffusionforcing} \\
                    \textcolor[HTML]{F4A700}{\rule{6pt}{6pt}} &MaskFlow (\textit{Ours})
                \end{tabular}
            };
    \end{tikzpicture}
    \vspace{-7pt}
    \caption{\textbf{Our method (MaskFlow) improves video quality compared to baselines while simultaneously requiring fewer function evaluations (NFE)} when generating videos $2\times$, $5\times$, and $10\times$ longer than the training window.
}
    \label{fig:teaser}
    \vspace{-10pt}
\end{figure}

\section{Introduction}

Due to the high computational demands of both training and sampling processes, long video generation remains a challenging task in computer vision. Many recent state-of-the-art video generation approaches train on fixed sequence lengths \cite{blattmann2023stable,blattmann2023align_videoldm,ho2022video} and thus struggle to scale to longer sampling horizons. Many use cases not only require long video generation, but also require the ability to generate videos with varying length. A common way to address this is by adopting an autoregressive diffusion approach similar to LLMs \cite{gao2024vid}, where videos are generated frame by frame. This has other downsides, since it requires traversing the entire denoising chain for every frame individually, which is computationally expensive. Since autoregressive models condition the generative process recursively on previously generated frames, error accumulation, specifically when rolling out to videos longer than the training videos, is another challenge.
\par
Several recent works \cite{ruhe2024rollingdiffusionmodels, chen2024diffusionforcing} have attempted to unify the flexibility of autoregressive generation approaches with the advantages of full sequence generation. These approaches are built on the intuition that the data corruption process in diffusion models can serve as an intermediary for injecting temporal inductive bias. Progressively increasing noise schedules \cite{xie2024progressive,ruhe2024rollingdiffusionmodels} are an example of a sampling schedule enabled by this paradigm. These works impose monotonically increasing noise schedules w.r.t. frame position in the window during training, limiting their flexibility in interpolating between fully autoregressive, frame-by-frame generation and full-sequence generation. This is alleviated in \cite{chen2024diffusionforcing}, where independent, uniformly sampled noise levels are applied to frames during training, and the diffusion model is trained to denoise arbitrary sequences of noisy frames. All of these works use continuous representations.
\par
\begin{figure}
    \centering
    \includegraphics[width=0.75\linewidth]{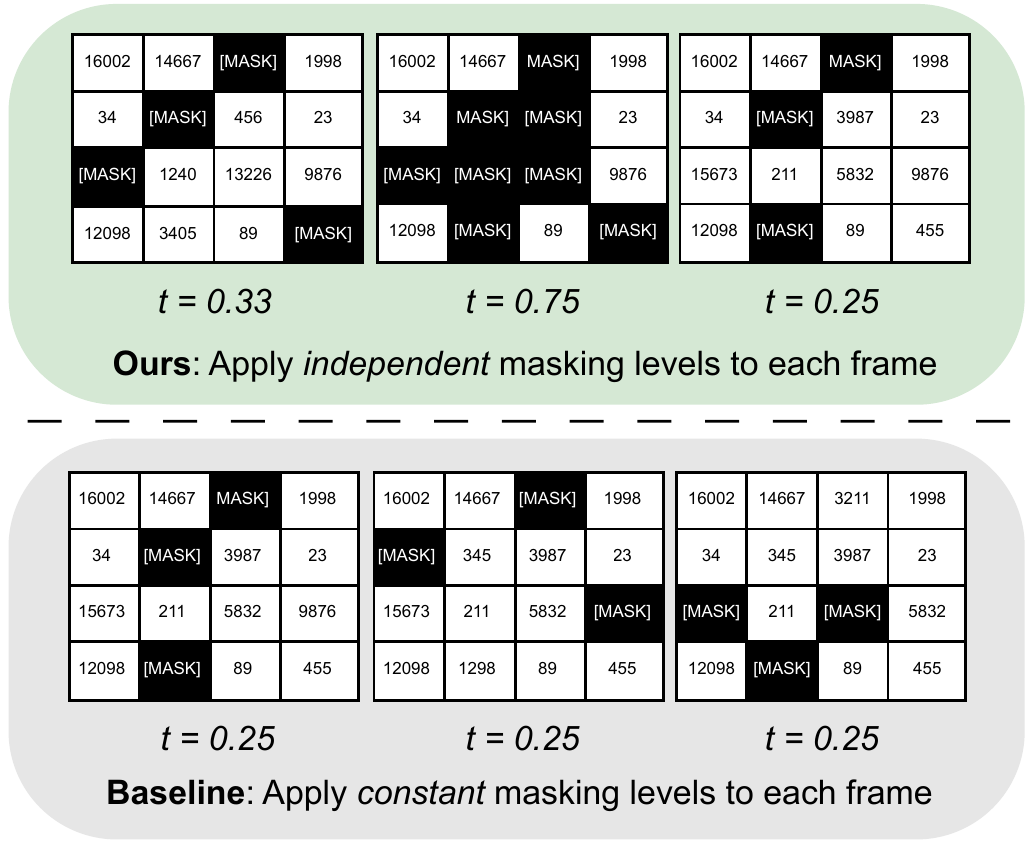}
    \caption{\textbf{MaskFlow Training:} For each video, Baseline training applies a single masking ratios to all frames, whereas our method samples masking ratios independently for each frame.}
    \vspace{-10pt}
    \label{fig:training}
\end{figure}

We transfer this idea to a discrete token space for two main reasons: First, it allows us to use a masking-based data corruption process, which enables confidence-based heuristic sampling that drastically speeds up the generative process. This becomes especially relevant when considering frame-by-frame autoregressive generation. Second, it allows us to use discrete flow matching dynamics, which provide a more flexible design space and the ability to further increase our sampling speed. Specifically, we adopt a \emph{frame-level masking} scheme in training (versus a \emph{constant-level masking} baseline, see Figure~\ref{fig:training}), which allows us to condition on an arbitrary number of previously generated frames while still being consistent with the training task. This makes our method inherently versatile, allowing us to generate videos using both full-sequence and autoregressive frame-by-frame generation, and use different sampling modes. We show that confidence-based masked generative model (MGM) style sampling is uniquely suited to this setting, generating high-quality results with a low number of function evaluations (NFE), and does not degrade quality compared to diffusion-like flow matching (FM)-style sampling that uses larger NFE. 
Combining frame-level masking during training with MGM-style sampling enables highly efficient long-horizon rollouts of our video generation models beyond $10 \times$ training frame lengths without degradation. We also demonstrate that this sampling method can be applied in a timestep-\emph{independent} setting that omits explicit timestep conditioning, even when models were trained in a timestep-dependent manner, which further underlines the flexibility of our approach. In summary, our contributions are the following:

\begin{itemize}
    \item To the best of our knowledge, we are the first to unify the paradigms of discrete representations in video flow matching with rolling out generative models to generate arbitrary-length videos. 
    \item We introduce MaskFlow, a frame-level masking approach that supports highly flexible sampling methods in a single unified model architecture.
    \item We demonstrate that MaskFlow with MGM-style sampling generates long videos faster while simultaneously preserving high visual quality (as shown in Figure~\ref{fig:teaser}).
    \item Additionally, we demonstrate an additional increase in quality when using full autoregressive generation or partial context guidance combined with MaskFlow for very long sampling horizons.
    \item We show that we can apply MaskFlow to both timestep-dependent and timestep-independent model backbones without re-training.
\end{itemize}

\section{Related Work}

\paragraph{Long Video Generation.} 
The training dynamics and the sampling methodology in this work are inspired by works like Diffusion Forcing \cite{chen2024diffusionforcing,song2025historyguidedvideodiffusion}, Rolling Diffusion Models \cite{ruhe2024rollingdiffusionmodels} and AR-Diffusion \cite{wu2023ar}. The main motivation behind these works is to unify the benefits of autoregression and full sequence diffusion by applying token-specific noise levels during training, which allows the model to generate future frames without fully denoising past frames in a sequence. \citet{xie2024progressive} is a similar work that prescribes a progressive sampling schedule for increased smoothness of transitions between generation windows. FIFO-Diffusion is a training-free inference approach for infinite text-to-video generation that uses a similar progressive denoising schedule and latent partitioning to reduce the training-inference gap with pre-trained video diffusion models. Other methods like \cite{gao2024vid, zheng2024open} and \cite{blattmann2023stable} use context frame conditioning similar to our method, but do not focus on long video generation. The closest to our work is \citet{zhou2025taming}, who also employ a masking-based design to generate arbitrary-length videos autoregressively. There are two key differences in our approach: We do not condition frame generation on any previous ground truth frames during training, but adopt a frame-level masking approach that is more flexible. We also employ confidence-based MGM-style sampling, which lets us sample entire training windows in very few sampling steps, whereas \citet{zhou2025taming} employs MAR-style \cite{li2024autoregressive} sampling that requires a higher amount of sampling steps per individual frame and does not use vector quantization.
\vspace{-10pt}

\paragraph{Discrete Representations in Video Generation.}
There are several previous works that investigate the use of discrete representations for video diffusion. MaskGIT \cite{chang2022maskgit} is a generative transformer that uses a bidirectional transformer decoder to predict randomly masked tokens in an input sequence of image patches. This idea is extended to videos in MAGVIT \cite{yu2023magvit}, which tokenizes video pixel space inputs into spatial-temporal visual tokens and uses a masked auto-regressive approach to predict masked input tokens. Similar approaches like Muse \cite{Chang2023MuseTG} and MAGVIT-v2\cite{yu2023language_magvit2} have shown promise in scaling up image and video generation tasks, but suffer from training instabilities. Latte \cite{ma2024latte} is a latent diffusion transformer model that uses a pre-trained VAE-based tokenizer to reduce the dimensions of frame sequences as well as a mixture of spatial and temporal attention blocks designed to decompose spatial and temporal dimensions of input sequences. We adapt this backbone to handle frame-level timestep conditioning to denoise frame sequences with independent masking levels. Unlike previous discrete methods~\cite{hu2024maskneed,ma2024latte} that do not explicitly consider frame dependence in the noise schedule, we investigate how combining multiple sampling styles and leveraging guidance from previously generated frames can yield an efficient and flexible long-video generation paradigm.

\paragraph{Discrete Flow Matching.}
Flow matching \cite{lipman2022flow} is an emerging generative modeling paradigm that generalizes common formulations of diffusion models and offers more freedom in the choice of the source distribution. Flow matching models have seen wide adoption in speech \cite{liu2023generative}, image generation \cite{hu2024zigma,hulfm,dao2023flow, lipman2022flow}, super-resolution \cite{schusterbauer2024boosting}, depth estimation \cite{gui2024depthfm} and video generation \cite{jin2024pyramidal}, but their application in high-dimensional discrete domains is still limited. Discrete flow matching \cite{gat2024discrete,campbell2024generative,shi2024simplifiedgeneralizedmaskeddiffusion,sahoo2024simpleeffectivemaskeddiffusion} addresses this limitation, introducing a novel discrete flow paradigm designed for discrete data generation. Building on this, \citet{hu2024maskneed} validates the efficacy of discrete flow matching in the image domain and bridges the connection between Discrete Diffusion and Masked Generative Models \cite{chang2022maskgit}. %
In contrast, we explore vectorizing timesteps across frames for memory-efficient long-video generation with improved extrapolation to long sampling horizons while also analyzing the impact of sampling styles on video quality.

\section{Method}

\begin{figure*}[ht!]
    \centering
    \includegraphics[width=0.9\linewidth]{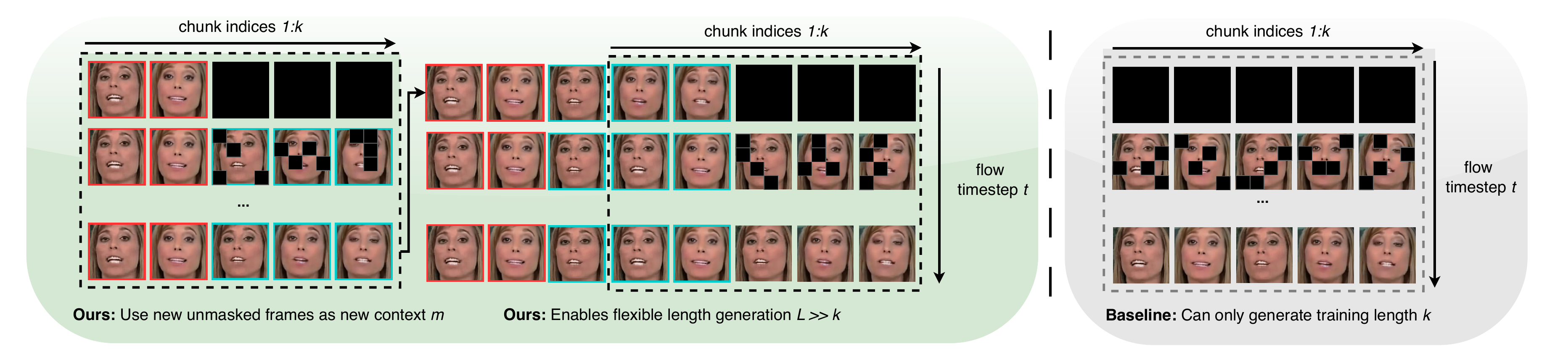}
    \vspace{-7pt}
    \caption{\textbf{MaskFlow Sampling:} Given $m=2$ \textcolor{red}{context frames} used to initialize generation, we unmask the current window and use \textcolor{color_generated_frame}{newly generated frames} as new context frames in the next chunk of size $k=5$, using stride $s=3$. (\textit{Tokenization omitted here to simplify understanding}) .
    }
    \label{fig:sampling}
    \vspace{-10pt}
\end{figure*}

\subsection{Task formulation: Long video generation}
\label{subsec:long_vid}

 There are, generally, three distinct approaches to long video generation. The first is the naive approach of training on long video sequences. This is challenging due to the quadratic complexity in attention mechanisms with respect to token numbers. Although works like\cite{tan2024video,harvey2022flexible} address this by distributing the generative process or by generating every \(n\)-th frame and subsequently infilling the remaining frames, the approach remains fundamentally resource-intensive. The second approach is a \emph{rolling} (or ``sliding-window'') approach, which applies monotonically increasing noise dependent on a frame's position in the sliding window. This process can be rolled out indefinitely, removing frames from the window when they are fully denoised and appending random noise frames at the end of the window. Works such as \cite{ruhe2024rollingdiffusionmodels, wu2023ar, xie2024progressive} belong to this paradigm. The third approach is \emph{chunkwise-autoregression}, also referred to as blockwise-autoregression \cite{ruhe2024rollingdiffusionmodels}. Here, the video of length $L$ is divided into overlapping \emph{chunks} of length $k \ll L$, where each chunk overlaps by $m$ frames, which we refer to as context frames. Concretely, we define a video and its frames as

\begin{equation}
   \mathbf{v} = (v^1, v^2, \dots, v^L)
\end{equation}

which we divide into overlapping chunks of length $k$. Let $\ell\;=\;\left\lceil \frac{L - k}{s} \right\rceil + 1$ denote the number of chunks needed to cover the video of length $L$, and we further define each chunk $\mathbf{v}^{(i)}$ as

\begin{equation}
   \mathbf{v}^{(i)} = \bigl(v^{(i-1)\,s + 1}, \dots, v^{(i-1)\,s + k}\bigr),
\end{equation}

where $s \le k$ is the sampling window stride, i.e., how far the context start shifts at each step. Often, one sets $s = k - m$, but this is not strictly required. The video distribution then factorizes as

\begin{equation}
\label{eq:markov}
   p(\mathbf{v};{\theta})
   \;=\;
   p(\mathbf{v}^{(1)};\theta)
   \prod_{i=2}^{\ell}
      p
         \bigl(
            \mathbf{v}^{(i)} 
            \;\bigm|\;
            \mathbf{v}^{(i-1)};\theta
         \bigr).
\end{equation}

Because each $\mathbf{v}^{(i)}$ overlaps the previous chunk by $m$ frames, the context frames feed into the next chunk's generation, ensuring smooth transitions and continuity between chunks. To enable such Markovian temporal dependencies during sampling, it is crucial to train a flexible backbone model \(p(\mathbf{v};\theta)\) that can generalize across different sampling schemes, such as the one defined in Equation~\eqref{eq:markov}.

\subsection{Preliminary: Flow Matching for Videos}
\label{subsec:flow}

Our masking flow matching approach, named \emph{MaskFlow}, draws inspiration from previous works that apply individual noise levels to individual frames in a sequence \cite{chen2024diffusionforcing, ruhe2024rollingdiffusionmodels}. These works operate in a continuous space, and use diffusion processes to corrupt data. MaskFlow operates in a discrete token space and uses \emph{masking} to corrupt data. We seek to learn a continuous transition process in ``time'' \(t\) that moves from a purely masked sequence at \(t=0\) to the unmasked token sequence at \(t=1\). In our method, the timestep $t$ corresponds to the masking ratio, and represents the frame-level probability of a token being masked. Consider a video consisting of \(L\) frames, where each frame is mapped to a discrete latent space using a vector-quantized (VQ) tokenizer \cite{esser2021taming_vqgan}. This tokenizer encodes each frame in the video $\mathbf{v}$ to a set of discrete latent indices $\mathbf{x}_{\text{latent}} \in [K]^N$, which consists of $N$ tokens drawn from the tokenizer vocabulary of size $K$. Let \(\mathcal{F}\) denote the VQ encoder-decoder, i.e., the function that maps a video in pixel space to its tokenized representation. Then, we have

\begin{equation}
    \mathbf{x} = \mathcal{F}(\mathbf{v}) \in [K]^{L \times N},
\end{equation}

where \([K] = \{1, 2, \ldots, K\}\) is the set of all possible token indices which includes a special ``mask token" $M \in [K]$. The choice of tokenization is essential here, since it compresses spatial dimensions of $\mathbf{x}$ compared to $\mathbf{v}$ and allows us to employ discrete flow matching, which we outline in further detail in the following section.

\begin{algorithm}[!ht]
\caption{\textbf{Training with Frame-level Masking}}
\label{alg:training}
\begin{algorithmic}[1]
\REQUIRE 
  Dataset of tokenized video clips $\mathcal{D}$, 
  network $p(\mathbf{x}_1 \mid \mathbf{x}_t, \mathbf{t};\theta)$, 
  chunk size $k$

\WHILE{not converged}
    \STATE \textbf{Sample} a chunk of $k$ frames  from $\mathcal{D}$, denoted 
      $\mathbf{x}_1 = (x_1^1, x_1^2, \dots, x_1^k)$
    %\STATE \textbf{Initialize} fully-masked chunk: 
      %$\mathbf{x}_0 = (\, [M], [M], \dots, [M] \,)$
    \FOR{$f = 1, \dots, k$}
        \STATE $t_f \sim \mathcal{U}(0,1)$
        \STATE 
           $x_{t^f} \;\sim\; p_{t^f \mid 0,1}\bigl(\,\cdot \mid x_0^f,\,x_1^f\bigr)$,
        where $p_{t^f \mid 0,1}$ follows 
        $
           (1 - t^f)\,\delta_{x_0^f} 
           \;+\; 
           t^f\,\delta_{x_1^f}.
        $
    \ENDFOR
    \STATE $\mathbf{x}_t = (x_{t^1}^1,\, x_{t^2}^2,\, \dots,\, x_{t^k}^k)$
    \STATE 
    $
       \hat{\mathbf{x}}_1 \;=\; p\bigl(\mathbf{x}_1 \mid \mathbf{x}_t,\, \mathbf{t};\theta\bigr),
    $
    where $\mathbf{t} = (t^1,\ldots,t^k)$
    \STATE \textbf{Backpropagate} $\mathcal{L}_\theta(\mathbf{x}_1, \hat{\mathbf{x}}_1)$ and \textbf{update} $\theta$.
\ENDWHILE
\end{algorithmic}
\end{algorithm}
\vspace{-10pt}

\paragraph{Discrete Flow Matching.}  
Discrete flow matching \cite{gat2024discrete} defines a vector field \(u_t\) in a discrete space that can be traversed to yield a smooth probability transition between our source distribution of fully masked frame sequences $p(\mathbf{x}_0)$ and the distribution of unmasked sequences $p(\mathbf{x}_1)$. This vector field defines an optimal transport path between the two distributions. Concretely, we construct the conditional probability path:
\begin{equation}
    p_{t \,\vert\, 0,1}\bigl(\mathbf{x} \,\vert\, \mathbf{x}_0, \mathbf{x}_1\bigr)
    \;=\; (1-t)\,\delta_{\mathbf{x}_0}(\mathbf{x})
    \;+\; t\,\delta_{\mathbf{x}_1}(\mathbf{x}),
\end{equation}

where \(\delta_{\mathbf{x}_0}(\mathbf{x})\) and \(\delta_{\mathbf{x}_1}(\mathbf{x})\) are Dirac delta functions (analogous to one hot encodings) in the discrete space that allocate all probability mass to the fully masked and fully unmasked sequences at $t=0$ and $t=1$, respectively. For any intermediate value \(t \in (0,1)\), the interpolation governed by the weights \((1-t)\) and \(t\) yields a new video sequence \(\mathbf{x}_t\) that represents a partially corrupted sequence. This is achieved by sampling each token from a mixture distribution where $1-t$ represents the probability of a token being masked.
\vspace{-10pt}
\paragraph{Kolmogorov Equation in Discrete State Spaces.}
In continuous-state models, one leverages the Continuity Equation \cite{song2021scorebased_sde} to ensure that a vector field \(u(\mathbf{x}_t, t)\) induces the desired probability transition between \(p(\mathbf{x}_0)\) and \(p(\mathbf{x}_1)\). The discrete counterpart is given by the Kolmogorov Equation \cite{campbell2024generative}, which similarly characterizes how a probability distribution evolves in time over discrete states. To achieve a transition between the fully masked and fully unmasked video distributions, we define the vector field:

\begin{equation}
    u_t(\mathbf{x}_t) 
    \;=\; 
    \frac{t}{\,1 - t} 
    \Bigl[
    p_{1 \,\vert\, t}(\mathbf{x}_1 \mid \mathbf{x}_t, t; \theta) 
    \;-\; 
    \delta_{\mathbf{x}_t}(x)
    \Bigr],
\end{equation}

where \(p_{1|t}(\mathbf{x}_1 \mid \mathbf{x}_t, t; \theta)\) is the model-predicted distribution of clean tokens \(\mathbf{x}_1\) given a partially corrupted sequence \(\mathbf{x}_t\) at time \(t\). Here, \(\delta_{\mathbf{x}_t}(x)\) represents the discrete Dirac delta centered at \(\mathbf{x}_t\). By following \(u_t\) through time, we recover a path that transforms \(p(\mathbf{x}_0)\) into \(p(\mathbf{x}_1)\).

\subsection{Training with Frame-Level Masking}
\label{subsec:training}

The flow matching formulation introduced in Sections~\ref{subsec:long_vid} and \ref{subsec:flow} employs a single scalar timestep $t$ to interpolate between the fully masked and fully unmasked video distributions. Our training procedure uses a reparametrization of this timestep. In our method, videos are generated in chunks, and only a subset of the frames (the non-context frames) are sampled from a fully masked initial state. To better simulate this process during training, we reparametrize the global timestep $t$ into a per-frame timestep vector $\mathbf{t}=(t^1,\dots, t^k)$ where each timestep $t^f$ specifies the masking ratio applied to frame $f$. In our setup, the context frames are assigned $t^f = 1$ (i.e. fully unmasked) while the new frames receive a masking level sampled from $\mathcal{U}(0,1)$.
By training the model to unmask frames with varying masking ratios per frame, we ensure that the network can effectively handle unmasked context frames while still learning a continuous transition from $p(\mathbf{x}_0)$ to $p(\mathbf{x}_1)$. To emphasize the reconstruction of masked tokens, we follow \cite{hu2024maskneed} in applying a masking operation on the cross-entropy loss. This results in the following objective:
\begin{multline}
\label{eq:ce_loss_masked}
    \mathcal{L}_{\theta} \;=\; 
    \E_{p(\mathbf{x}_1)\,p(\mathbf{x}_0)\,\mathcal{U}(\mathbf{t};0,1)\,p_{t|0,1}(\mathbf{x}_t \,\vert\, \mathbf{x}_0,\mathbf{x}_1)} \\
    \Bigl[\;\underbrace{\delta_{[M]}(\mathbf{x}_t)\,( \mathbf{x}_1 )^\top}_{\text{Loss Masking}}
    \;\log \denoise(\mathbf{x}_1 \,\vert\, \mathbf{x}_t,\mathbf{t};\theta) 
    \Bigr],
\end{multline}

where $\delta_{[M]}(\mathbf{x}_t)$ indicates that only masked tokens are used in the cross-entropy computation. The choice of frame-level masking is essential because it aligns the task of generating chunks of size $k$ conditioned on $m$ clean context frames with our training task. In both scenarios, our models are tasked with unmasking frame sequences with varying masking levels across frames. We show that compared to a constant masking level baseline, this training choice enables chunkwise autoregressive rollout to long sequence lengths. Our training algorithm is shown in detail in Algorithm~\ref{alg:training}.

\subsection{Chunkwise Autoregression for Long Videos}
\label{sec:flex_long}

 To generate a coherent video of length \(L \gg k\), we employ the chunkwise autoregressive approach as described previously. Let \(m\) be the number of context frames provided to the model (drawn initially from ground-truth, later from previous generated frames). In each iteration, we pass \(k\) frames to the model, where the first \(m\) of these frames are context and the remaining \((k - m)\) frames are fully masked. The model unmasks these frames. Afterwards, we shift the context window forward by \(s\) and repeat this process, until we have generated \(L\) total frames. Figure~\ref{fig:sampling} illustrates this pipeline. Note that we dynamically increase the number of context frames $m$ in the final chunk in case there are less than $s$ frames left to generate. In those cases we set $m = k - R$ where $R$ is the remaining number of frames, giving the final chunk a larger context. We do this to avoid generating video lengths beyond $L$ which would result in either discarding generated frames or generating videos longer than $L$. This is shown in detail in Algorithm~\ref{alg:chunkwise}.

\paragraph{Autoregressive \textit{v.s.} Full-Sequence Generation.}
By varying the stride \(s\), we can interpolate between (i) a fully autoregressive mode (\(s = 1\)) with $m = k - s$, where we generate a single new frame per chunk, and (ii) a full-sequence mode (\(s = k - m\)), where we generate $k - m$ new frames simultaneously in each chunk. Smaller \(s\) increases compute cost but may yield higher frame quality, whereas larger \(s\) is more efficient, but may result in a drop in frame quality. Our experimental results shown in Table~\ref{tab:autoregression} support this intuition.

\paragraph{FM-Style \textit{v.s.} MGM-Style Sampling.}
MaskFlow supports two distinct sampling modes. In FM-style sampling, we gradually traverse the probability path from the fully masked sequence $\mathbf{x}_0$ to the final unmasked sequence $\mathbf{x}_1$. A smaller step size yields smoother transitions at the cost of more denoising steps. Alternatively, in MGM-style sampling, we apply confidence-based heuristic sampling similar to \citet{chang2022maskgit}. In each sampling step, the model computes token-wise confidence scores for each predicted token and selects a fraction of the most confident tokens to unmask. This sampling process allows us to generate video chunks efficiently in much fewer sampling steps.
\vspace{-10pt}

\paragraph{Timestep-dependent models and timestep-independent sampling.}

By default, our model backbones are timstep-dependent, meaning each forward pass receives a timestep vector $\mathbf{t}\in[0,1]^k$ that indicates the masking ratio of each frame. Internally, we embed $\mathbf{t}$ through a learnable mapping to produce conditioning vectors that modulate various layers (e.g., via layer norm shifts/scales). Interestingly, we can still sample these models timstep-independently. Concretely, when using MGM-style sampling, we iteratively unmask a chunk of tokens while simply passing $\mathbf{t}=\mathbf{0}$ at each iteration, effectively treating our timestep-dependent model as if it were timestep-independent:

\begin{equation}
    p(\mathbf{x}_1|\mathbf{x}_t;\theta) \approx p(\mathbf{x}_1|\mathbf{x}_{t}, \mathbf{t}=\mathbf{0};\theta).
\end{equation}

This works, since the learned network can infer the corruption state (mask ratio) from the input tokens alone. Thus, in practice, \emph{a single} trained model can serve both as a standard time-dependent (flow-matching) generator \emph{and} as a time-independent (MGM-style) sampler, providing greater flexibility at inference time.

\begin{algorithm}[ht!]
\caption{Chunkwise Autoregression for Long Videos}
\label{alg:chunkwise}
\begin{algorithmic}[1]
\REQUIRE Video length \(L\), context frames \(\mathbf{x}^{1:m} = (x^1,\dots,x^m)\), chunk size \(k\), stride \(s\), fully masked frame \([M]\), network \(p(\mathbf{x}_1 \mid \mathbf{x}_t,\mathbf{t};\theta)\)
\STATE \textbf{Initialize:} \(\hat{\mathbf{x}}_{1} \leftarrow (x^1,\dots,x^m)\); \(c \leftarrow m\) \COMMENT{current frame}
\WHILE{\(c < L\)}
    \STATE \(R \leftarrow L - c\) \COMMENT{remaining frames}
    \STATE \(h \leftarrow \min(R,\, s)\) \COMMENT{frames to generate this chunk}
    \IF{\(R \le s\)}
        \STATE \(m \leftarrow k - R\)
    \ENDIF
    \STATE \(\mathbf{x}_{\mathrm{context}} \leftarrow (x^{\,c-m+1}, \dots, x^c)\)
    \STATE \(\mathbf{x}_{\mathrm{mask}} \leftarrow (\underbrace{[M], \dots, [M]}_{h\text{ times}})\)
    \STATE \(\mathbf{x}_{\mathrm{out}} \sim p\Bigl(\mathbf{x}_1 \mid (\mathbf{x}_{\mathrm{context}},\, \mathbf{x}_{\mathrm{mask}}), \mathbf{t};\theta\Bigr)\)
    \STATE \(\mathbf{x}_{\mathrm{new}} \leftarrow (x_{\mathrm{out}}^{\,m+1}, \dots, x_{\mathrm{out}}^{\,m+h})\)
    \STATE \(\hat{\mathbf{x}}_1 \leftarrow (\hat{\mathbf{x}}_1,\, \mathbf{x}_{\mathrm{new}})\)
    \STATE \(c \leftarrow c + h\)
\ENDWHILE
\RETURN \(\hat{\mathbf{x}}_1\)
\end{algorithmic}
\end{algorithm}

\vspace{-1pt}
\section{Experiments}

\subsection{Datasets and Evaluation Metrics}

\paragraph{Datasets.} We mainly consider two datasets: Deepmind Lab (DMLab) for evaluating performance in diverse ego-centric views and FaceForensics (FFS) for assessing video fluency. DMLab contains videos of random walks in a 3D maze, while FFS consists of deepfakes. Both datasets are preprocessed and tokenized using SD-VQGAN \cite{rombach2022high_latentdiffusion_ldm} for training. Further details are provided in the Appendix.

\paragraph{Evaluation metrics: FVD for video quality, NFE for sampling efficiency.}  For video generation, we use Fréchet Video Distance (FVD) \cite{unterthiner2018towards} as our main evaluation metric. For FVD, we adhere to the evaluation guidelines introduced by StyleGAN-V \cite{ma2024latte,skorokhodov2022stylegan_v}.
For all generation experiments requiring context frames, we randomly sample consecutive context frames from each ground-truth video in the dataset, and generate a corresponding generated video using our trained models. 
To compute FVD, we use a randomly sampled window of $L$ frames from the ground-truth videos, and sample the same number of generated videos using our models. This amounts to 704 videos for FFS, and 625 videos for DMLab FVD calculation across different sampling horizons $L$. 
We additionally evaluate the sampling efficiency of our method against various baselines by comparing the required number of function evaluations (NFE) and sampling wall clock times using identical compute resources.

\subsection{Training details}

We use a vocabulary size $K=16{,}384$ and token length $1{,}024$ to compress video frames by a compression factor of $8$. We then train on a small subset of training sequences of $k=16$ frames for FFS and $k=36$ frames for DMLab. We use a Latte XL2 \cite{ma2024latte} backbone with 760M parameters for all FFS experiments, and a smaller Latte B2 backbone architecture with 129M parameters for DMLab, and train it using discrete flow matching dynamics. Please refer to the Appendix for more detailed information about the training recipe and hyperparameters.

\subsection{Main Results}

\begin{table}[ht!]
    \centering
    \normalsize
    \resizebox{0.47\textwidth}{!}{
    \begin{tabular}{l|crr}
    \toprule
    \textbf{Sampling Mode} & \makecell{\textbf{Extrapolation} \\ \textbf{Factor}} 
    & \makecell{\textbf{Total} \\ \textbf{NFE}} 
    & \makecell{\textbf{FVD} $\downarrow$} \\
    \midrule
        Diffusion Forcing~\cite{chen2024diffusionforcing} & $2\times$ & $798$ & 144.43 \\ 
        Rolling Diffusion~\cite{ruhe2024rollingdiffusionmodels} & $2\times$ & $750$ & 72.49 \\
        \hline
        \textit{MaskFlow} (FM-Style) & $2\times$ & $788$ & $66.94$ \\
        \rowcolor{gray!8}\textit{MaskFlow} (MGM-Style) & $2\times$ & \textbf{60}  & \textbf{59.93} \\
        \midrule
        \midrule
        Diffusion Forcing~\cite{chen2024diffusionforcing} & $5\times$ & $1{,}596$ & 272.14 \\
        Rolling Diffusion~\cite{ruhe2024rollingdiffusionmodels} & $5\times$ & $1{,}652$ & 248.13 \\ 
        \hline
        \textit{MaskFlow} (FM-Style) & $5\times$ & $1{,}500$ & $118.81$ \\
        \rowcolor{gray!8}\textit{MaskFlow} (MGM-Style) & $5\times$ & \textbf{120}  & \textbf{108.74} \\
        \midrule
        \midrule
        Diffusion Forcing~\cite{chen2024diffusionforcing} & $10\times$ & $3{,}192$ & 306.31 \\
        Rolling Diffusion~\cite{ruhe2024rollingdiffusionmodels} & $10\times$ & $3{,}092$ & 451.38 \\ 
        \hline
        \textit{MaskFlow} (FM-Style) & $10\times$ & $3{,}000$ & \textbf{174.85} \\
        \rowcolor{gray!8}\textit{MaskFlow} (MGM-Style) & $10\times$ & \textbf{240}  & $214.39$ \\
        \bottomrule
    \end{tabular}}
    \vspace{-7pt}
    \caption{\textbf{Both MGM-style and FM-style sampling extrapolate to longer sequences with similar FVD, but MGM-style is much faster.} Performance deteriorates for larger extrapolation factors, but MaskFlow consistently outperforms Diffusion Forcing and Rolling Diffusion. Results are on timestep-dependent FaceForensics models with full sequence generation ($s = k - m$).}
    \label{tab:extrapolation}
\end{table}

\begin{comment}
\begin{figure}
    \centering
    \begin{subfigure}[t]{0.495\columnwidth}
        \centering
        \includegraphics[width=\linewidth]{figpaper/nfe_vs_fvd_vs_ep_ffs.pdf}
        \caption{\textbf{FaceForensics}}
        \label{fig:ffs}
    \end{subfigure}\hfill
    \begin{subfigure}[t]{0.495\columnwidth}
        \centering
        \includegraphics[width=\linewidth]{figpaper/nfe_vs_fvd_vs_ep_dmlab.pdf}
        \caption{\textbf{DMLab}}
        \label{fig:dmlab}
    \end{subfigure}
    \caption{\textbf{MaskFlow performance scales favorably across NFE for different extrapolation factors across both datasets.} Shows a comparison between MaskFlow full sequence and MaskFlow autoregressive modes and other baselines across extrapolation factors.}
    \vspace{-10pt}
    \label{fig:maskflow_scaling}
\end{figure}
\end{comment}

\begin{figure}
    \centering
    \includegraphics[width=\linewidth]{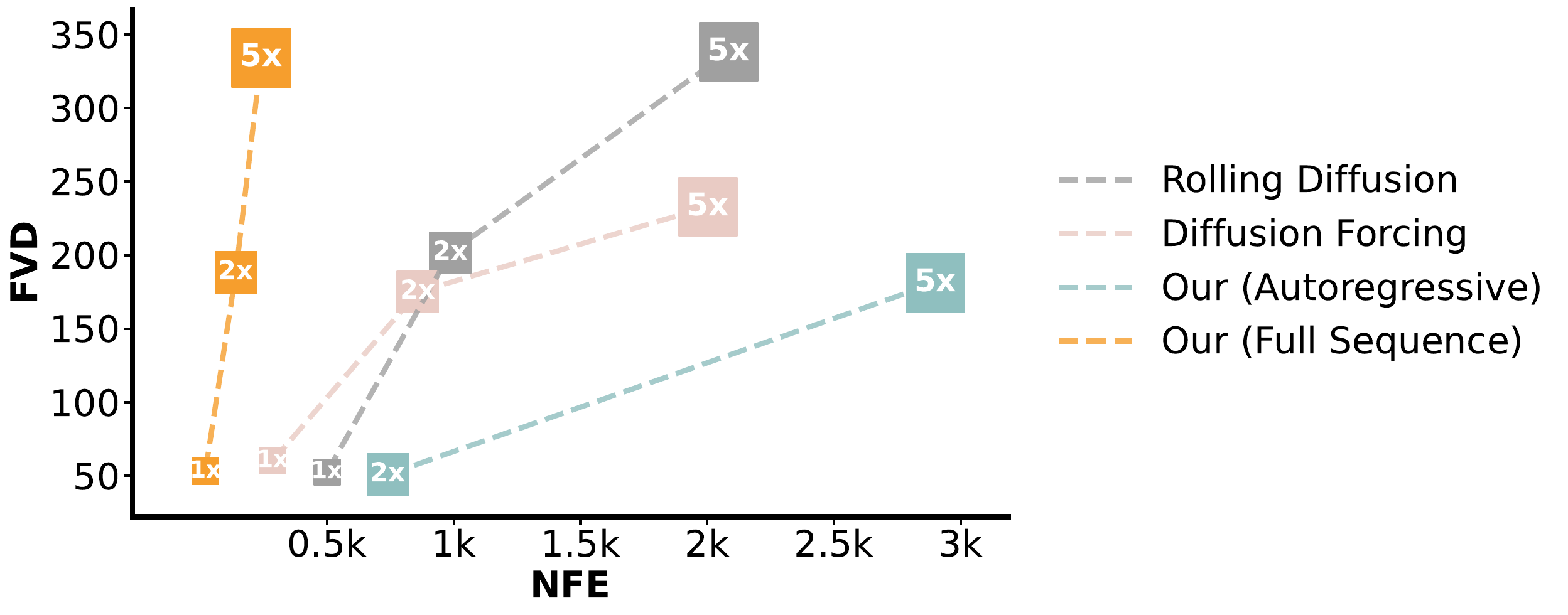}
    %\label{fig:dmlab}
    \vspace{-18pt}
    \caption{\textbf{MaskFlow performance scales favorably across NFE for different extrapolation factors.} Shows a comparison between MaskFlow full sequence and MaskFlow autoregressive modes and other baselines across extrapolation factors on DMLab.}
    \vspace{-10pt}
    \label{fig:maskflow_scaling}
\end{figure}

\paragraph{Baselines.}
The two most comparable works to our method are \citet{chen2024diffusionforcing} and \citet{ruhe2024rollingdiffusionmodels}. Both of these techniques propose novel sampling methods that can be rolled out to long video lengths, and also apply frame-specific noise levels. Both of these approaches are diffusion-based and operate on continuous representations, whereas we operate on discrete tokens and use masking. We re-implement both the pyramid sampling scheme proposed in Diffusion Forcing and the Rolling Diffusion sampling method in our discrete setting. This allows us to compare the baseline sampling methods to MaskFlow on the same model backbones. We also compare MaskFlow to a constant masking level baseline from \citet{hu2024maskneed} to evaluate the design choice of frame-level masking.

\begin{figure}
    \centering
    \includegraphics[width=1.0\columnwidth]{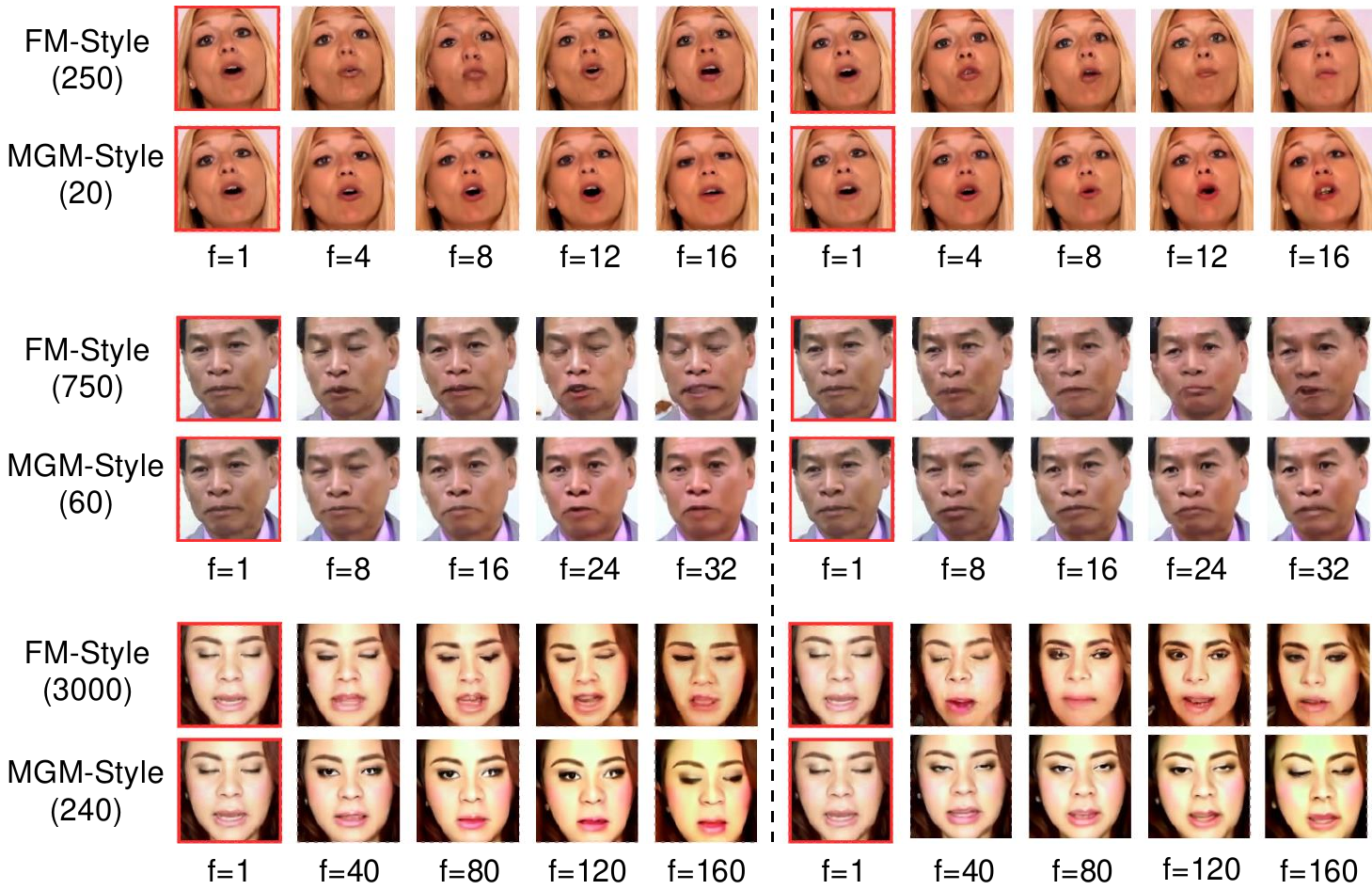}
    \vspace{-17pt}
    \caption{\textbf{MGM-style sampling generates visually pleasing videos with two \textcolor{red}{context frames} beyond $10\times$ training frame length with only 20 sampling steps.} Shows sampling mode and total NFE in brackets, and frame indices $f$. The left and right subfigures show distinct videos obtained with identical sampling modes and context frames.
    }
    \vspace{-10pt}
    \label{fig:faces1}
\end{figure}

\paragraph{Our MGM-style sampling approach can generate long videos efficiently with minimal degradation.} 
Table~\ref{tab:extrapolation} shows the ability of our model to generate long videos. We define the \emph{extrapolation factor} as the ratio of sampling and training window lengths, so an extrapolation factor of $2 \times$ means we generate videos twice as long as the training videos, e.g. $32$ frames for FFS on a training window size of $k=16$ frames. The experiments in Table~\ref{tab:longer_train_window_baseline} of the Appendix all use full sequence generation with $s = k - m$. While video quality deteriorates for longer extrapolation factors due to error accumulation, our method is able to maintain visual quality for large extrapolation factors. This ability is enabled by our training approach, which ensures that our models are able to unmask arbitrary mixtures of low and high masking ratio frames. This allows us to condition each chunk on arbitrary numbers of previously generated frames, which is consistent with the training task. A detailed qualitative overview is shown in Figure~\ref{fig:faces1}. Both FM-style and MGM-style sampling modes retain this ability, but our MGM-style sampling generates high-quality results with lower NFE. We also show that MaskFlow outperforms both Rolling Diffusion~\cite{ruhe2024rollingdiffusionmodels} and Diffusion Forcing~\cite{chen2024diffusionforcing} with pyramid noise schedule in discrete settings.

\paragraph{Frame-level masking does not reduce performance on original training window length generation.} Table~\ref{tab:baseline_training_window} shows that our frame-level masking approach does not reduce performance for a single chunk compared to a constant masking baseline. We compare a frame-level masking DMLab model trained on $k=36$ frames with a constant masking baseline and show that our frame-level masking models outperform the constant masking baseline across two sampling modes. This demonstrates that our frame-level masking training does not trade off quality on training window length generation for the ability to generate longer videos.

\begin{figure}[ht]
    \centering
    \includegraphics[width=1.0\linewidth]{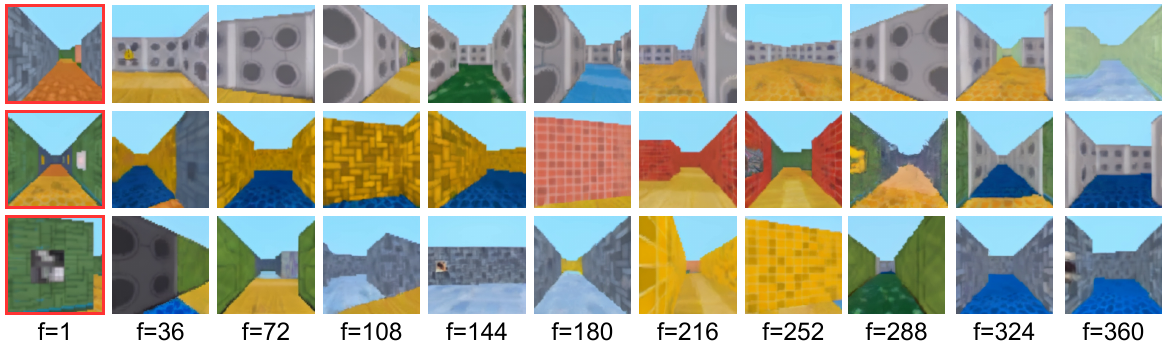}
    \vspace{-20pt}
    \caption{\textbf{Fully autoregressive sampling stabilizes DMLab videos beyond extrapolation factor $10 \times$.} All examples use fully autoregressive MaskFlow (MGM-style) sampling with $s=1$ and 6,500 NFE in total. The final context frame is shown in red.}
    \vspace{-10pt}
    \label{fig:dmlabauto}
\end{figure}

\paragraph{Fully Autoregressive Sampling increases video quality at the cost of inference speed.} To further illustrate the flexibility of our method, we run a series of experiments using a sampling stride of $s=1$ with $m = k-1$. We thus initialize the generative process by conditioning on almost a full training clip, and then generating new frames frame by frame using our existing sampling approaches. This requires us to traverse the entire unmasking chain for each generated frame, making this sampling method slower than the sampling approach employed in Table~\ref{tab:extrapolation}. Specifically on DMLab, which is more dynamic than FFS, this substantially improves results, enabling extremely long high-quality rollouts (see Figure~\ref{fig:dmlabauto}. The findings in Table ~\ref{tab:autoregression} thus demonstrate that for certain datasets, such as FFS, iterative full sequence generation already works very well, whereas autoregressive sampling is more suitable for more dynamic datasets, such as DMLab. Since our MGM-style sampling is able to generate new frames in very few NFE, autoregressive frame-by-frame generation actually requires a similar NFE than the baselines that do full sequence generation with FM-style sampling. Figure~\ref{fig:maskflow_scaling} highlights this, showing that MaskFlow scales favorably compared to other methods in terms of NFE for $s=1$ and $s=k-m$. A more detailed comparison of autoregressive and full sequence sampling in terms of wall clock sampling speed can be found in Table~\ref{tab:speed_comparison} of the Appendix.

\begin{table}[ht]
    \centering
    \normalsize
    \resizebox{0.47\textwidth}{!}{%
    \begin{tabular}{l|ccrr}
        \toprule
        & \makecell{\textbf{Extrapolation} \\ \textbf{Factor}}
        & \makecell{\textbf{Sampling} \\ \textbf{Stride}}
        & \makecell{\textbf{Total} \\ \textbf{NFE}}
        & \makecell{\textbf{FVD} $\downarrow$} \\
        \midrule
        FaceForensics   & $2\times$  & $s=14$ (\textit{full sequence}) & \textbf{60} & 59.93 \\
        FaceForensics   & $2\times$  & $s=1$ (\textit{autoregressive})  & 340 & \textbf{30.43} \\
        \midrule
        FaceForensics   & $5\times$  & $s=14$ (\textit{full sequence}) & \textbf{120} & 108.74 \\
        FaceForensics & $5\times$  & $s=1$ (\textit{autoregressive})  & 1,300 & \textbf{103.69} \\
        \midrule
        FaceForensics   & $10\times$ & $s=14$ (\textit{full sequence}) & \textbf{240} & 214.39 \\
        FaceForensics   & $10\times$ & $s=1$ (\textit{autoregressive})  & 2,900 & \textbf{165.02} \\
        \midrule
        \midrule
        DMLab & $2\times$  & $s=24$ (\textit{full sequence}) & \textbf{60} & 195.84 \\
        DMLab & $2\times$  & $s=1$ (\textit{autoregressive})  & 740  & \textbf{42.53} \\
        \midrule
        DMLab & $5\times$  & $s=24$ (\textit{full sequence}) & \textbf{140}  & 334.15 \\
        DMLab & $5\times$  & $s=1$ (\textit{autoregressive})  & 2,900 & \textbf{80.56} \\
        \bottomrule
    \end{tabular}
    }
    \vspace{-7pt}
    \caption{\textbf{Fully autoregressive sampling significantly improves performance on DMLab but also increases the required NFE.} Results are obtained using best-performing models with MGM-style sampling mode.}
     \vspace{-10pt}
    \label{tab:autoregression}
\end{table}

\begin{table}
    \centering
    \scriptsize
    \begin{tabular}{c c r}
        \toprule
          \multirow{2}{*}{\textbf{Extrapolation}} & \multirow{2}{*}{\textbf{Guidance}} 
          & \textbf{FVD} $\downarrow$ \\
        \cmidrule(lr){3-3}
          \textbf{Factor} & \textbf{Level $\omega$} 
          & DMLab \\
        \midrule
        $1\times$
        & $0$
        & \textbf{45.84} \\
        $1\times$
        & $1.0$
        & \text{49.76} \\
       $1\times$
       & $1.5$
       & \text{47.25} \\
       $1\times$
        & $2.0$
         & \text{46.29} \\
       \hline
        $2\times$
          & $0$
          & \text{219.33} \\
        $2\times$
          & $1.0$
          & \text{189.48} \\
        $2\times$
          & $1.5$
          & \text{167.80} \\
        \rowcolor{gray!8}$2\times$
          & $2.0$
          & \textbf{141.94} \\
        \hline
        $5\times$
          & $0$
          & \text{402.73} \\
        $5\times$
          & $1.0$
          & \text{403.32} \\
        $5\times$
          & $1.5$
          & \text{315.26} \\
        \rowcolor{gray!8} $5\times$
          & $2.0$
          & \textbf{281.20} \\
        \bottomrule
    \end{tabular}
    %}
    \vspace{-7pt}
    \caption{\textbf{Scaling partial context guidance $\omega$ can substantially improve performance for longer extrapolation factors.} Results use MaskFlow with MGM-Style sampling and $s = k - m$.}
    \vspace{-10pt}
    \label{tab:partial_context}
\end{table}

\paragraph{Scaling partial context guidance further improves performance on full sequence generation.}
Inspired by classifier-free guidance \cite{ho2021classifier} and history guidance in Diffusion Forcing \cite{chen2024diffusionforcing,song2025historyguidedvideodiffusion}, we propose a training-free sampling method that fuses multiple model predictions of \(\,p(x_{1}| x_{t};\theta)\) using different levels of conditioning on past frames. 
Concretely, we run forward passes where \(x_t\) contains: 
(i) \emph{no} context frames (unconditional) , 
(ii) \emph{partially masked} context frames (partial conditioning), 
and (iii) \emph{fully clean} context frames (fully conditional). 
We then fuse the predicted logits with a guidance scale $\omega$. 
By using \emph{partially masked} rather than fully clean context frames for some of these passes, the model is encouraged to preserve global movement and dynamics without strictly copying the observed context. Formally, if \(z_{\mathrm{uncond}} (i),\ z_{\mathrm{partial}}(ii),\ z_{\mathrm{cond}} (iii)\) denotes logits from the three forward passes, one can construct a composite logit distribution via
$z_{\mathrm{cond}} + \omega \cdot (z_{\mathrm{partial}} - z_{\mathrm{uncond}})$ that balances sample variety (unconditional) with temporal coherence (partial and full context). Partial context guidance requires no re-training and can yield improved fidelity and motion consistency. Table~\ref{tab:partial_context} shows performance improvements achieved on timestep-independent DMLab models.

\begin{table}[t]
    \centering
    \normalsize
    \resizebox{\columnwidth}{!}{%
    \begin{tabular}{l c c}
        \toprule
          \multirow{2}{*}{\textbf{Training}} & \multirow{2}{*}{\textbf{Sampling}} 
          & \textbf{FVD} $\downarrow$ \\
        \cmidrule(lr){3-3}
          \textbf{Mode} & \textbf{Mode (NFE)} 
          & DMLab \\
        \midrule
        Constant Masking \cite{hu2024maskneed}$^\dagger$ 
          & FM-Style
          & \text{53.31} \\
       
          Frame-level Masking 
          & Diffusion Forcing~\cite{chen2024diffusionforcing} 
          & 60.30 \\
          Frame-level Masking 
          & Rolling Diffusion~\cite{ruhe2024rollingdiffusionmodels}
          & 52.43 \\
          \hline 
          \hline 
        Frame-level Masking 
          & \textit{MaskFlow} (MGM-Style)  
          & 53.17 \\
           Frame-level Masking 
          & \textit{MaskFlow} (FM-Style)
          & \textbf{49.62} \\
        \bottomrule
    \end{tabular}
    }
    {\fontsize{6}{7}\selectfont ($\dagger$) denotes pretrained by us using their official implementation.}
    \vspace{-7pt}
    \caption{\textbf{Frame-level masking performs on par with constant masking when sampling window equals training window length videos}. MGM-style sampling performs well with only 20 NFE.}
    \label{tab:baseline_training_window}
\end{table}

\subsection{Ablations}

\begin{table}[t]
\centering
\normalsize
\resizebox{\columnwidth}{!}{
\begin{tabular}{cc|cccc}
\toprule
\multirow{2}{*}{\textbf{Sampling}} & \multirow{2}{*}{\textbf{Model}} & \multirow{2}{*}{\textbf{Sampling-}} & \multirow{2}{*}{\textbf{Extrap.}} & \multicolumn{2}{c}{\textbf{FVD} $\downarrow$} \\
\cmidrule(lr){5-6}
\textbf{Mode} & \textbf{Time Dep.} & \textbf{Time Indep.} & \textbf{Factor} & DMLab & FaceForensics \\
\midrule
FM-Style    & \textcolor{green}{\cmark} & \textcolor{red}{\xmark} & 1× & 55.19 & 48.98 \\
MGM-Style   & \textcolor{red}{\xmark} & \textcolor{green}{\cmark} & 1× & \textbf{45.84} & 77.04 \\
\rowcolor{gray!8}MGM-Style   & \textcolor{green}{\cmark} & \textcolor{red}{\textcolor{green}{\cmark}} & 1× & 53.17 & \textbf{45.92} \\
\midrule
FM-Style    & \textcolor{green}{\cmark} & \textcolor{red}{\xmark} & 2× & 267.80 & 66.94 \\
MGM-Style   & \textcolor{red}{\xmark} & \textcolor{green}{\cmark} & 2× & 219.33 & 109.96 \\
\rowcolor{gray!8}MGM-Style   & \textcolor{green}{\cmark} & \textcolor{green}{\cmark} & 2× & \textbf{188.22} & \textbf{59.93} \\
\midrule
FM-Style    & \textcolor{green}{\cmark} & \textcolor{red}{\xmark} & 5× & 360.61 & 118.81 \\
MGM-Style   & \textcolor{red}{\xmark} & \textcolor{green}{\cmark} & 5× & 402.73 & 137.66 \\
\rowcolor{gray!8}MGM-Style   & \textcolor{green}{\cmark} & \textcolor{green}{\cmark} & 5× & \textbf{334.15} & \textbf{108.74} \\
\bottomrule
\end{tabular}}
\vspace{-7pt}
\caption{\textbf{Timestep-dependent models can generate high-quality results with timestep-independent sampling.} Timestep-dependent models with timestep-independent sampling show best results across various extrapolation factors.}
\vspace{-10pt}
\label{tab:time_condition}
\end{table}

\paragraph{Timestep-dependent models can be sampled in a time-independent training-free manner.}
An additional interesting observation is that MGM-style sampling without explicit timestep conditioning is able to generate high-quality results in the full-sequence case. We thus compare timestep-dependent and timestep-independent models under different sampling modes in Table ~\ref{tab:time_condition}. Our results demonstrate that the timestep-dependent models when sampled with MGM-style sampling actually perform best. We hypothesize that this is due to the more explicit inductive bias of timestep conditioning during training, and that this guides the learning process towards improved unmasking irrespective of the actual timesteps passed during inference. We are thus able to apply our sampling modes across timestep-dependent and independent models without requiring any re-training, which further underlines the flexibility of our approach.
\vspace{-10pt}

\begin{figure}[ht!]
    \centering
    \begin{subfigure}[b]{0.495\columnwidth}
        \centering
        \includegraphics[width=\linewidth]{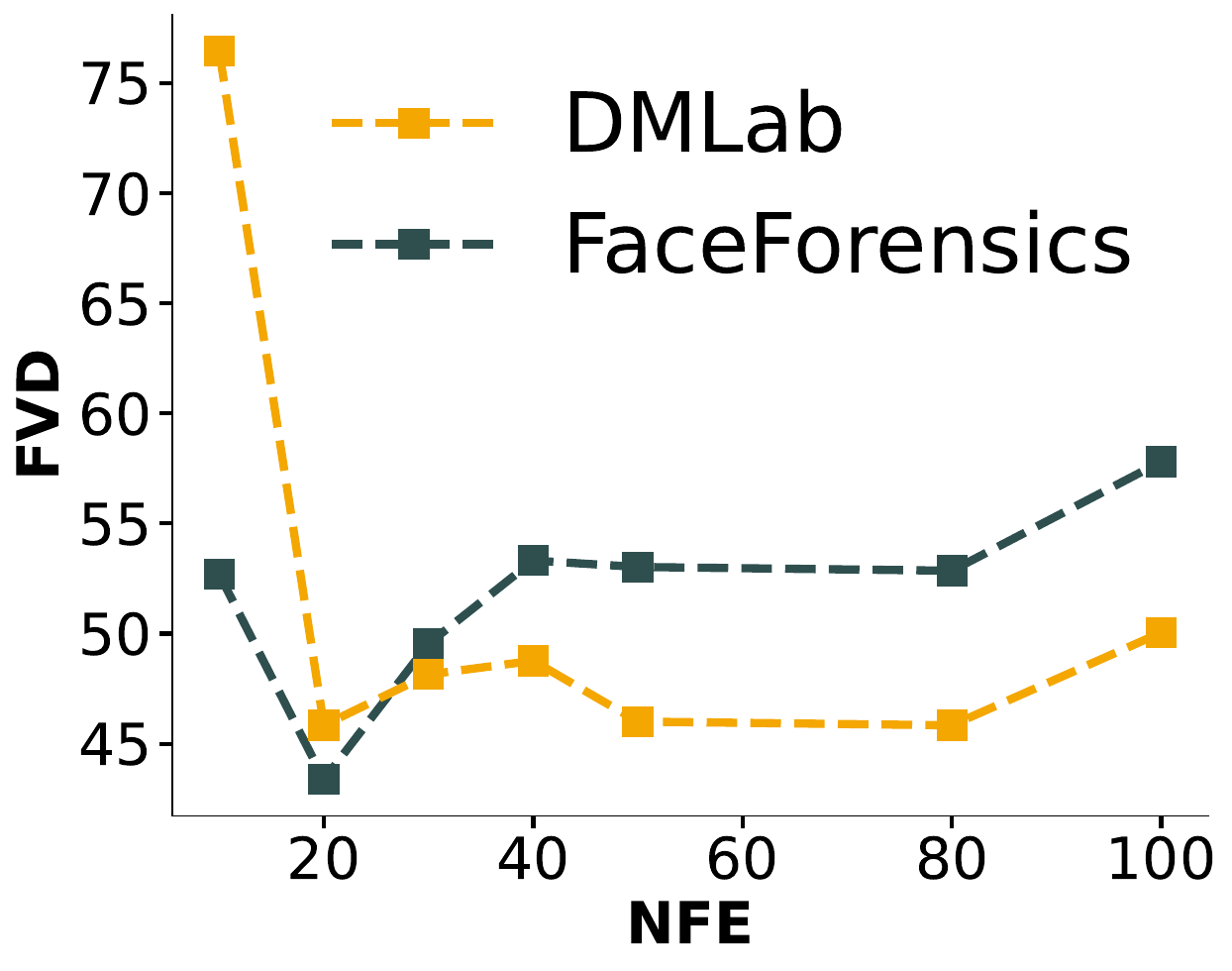}
        \caption{MGM-style sampling}
        \label{fig:nfe_fvd_mgm}
    \end{subfigure}
    \hfill
    \begin{subfigure}[b]{0.495\columnwidth}
        \centering
        \includegraphics[width=\linewidth]{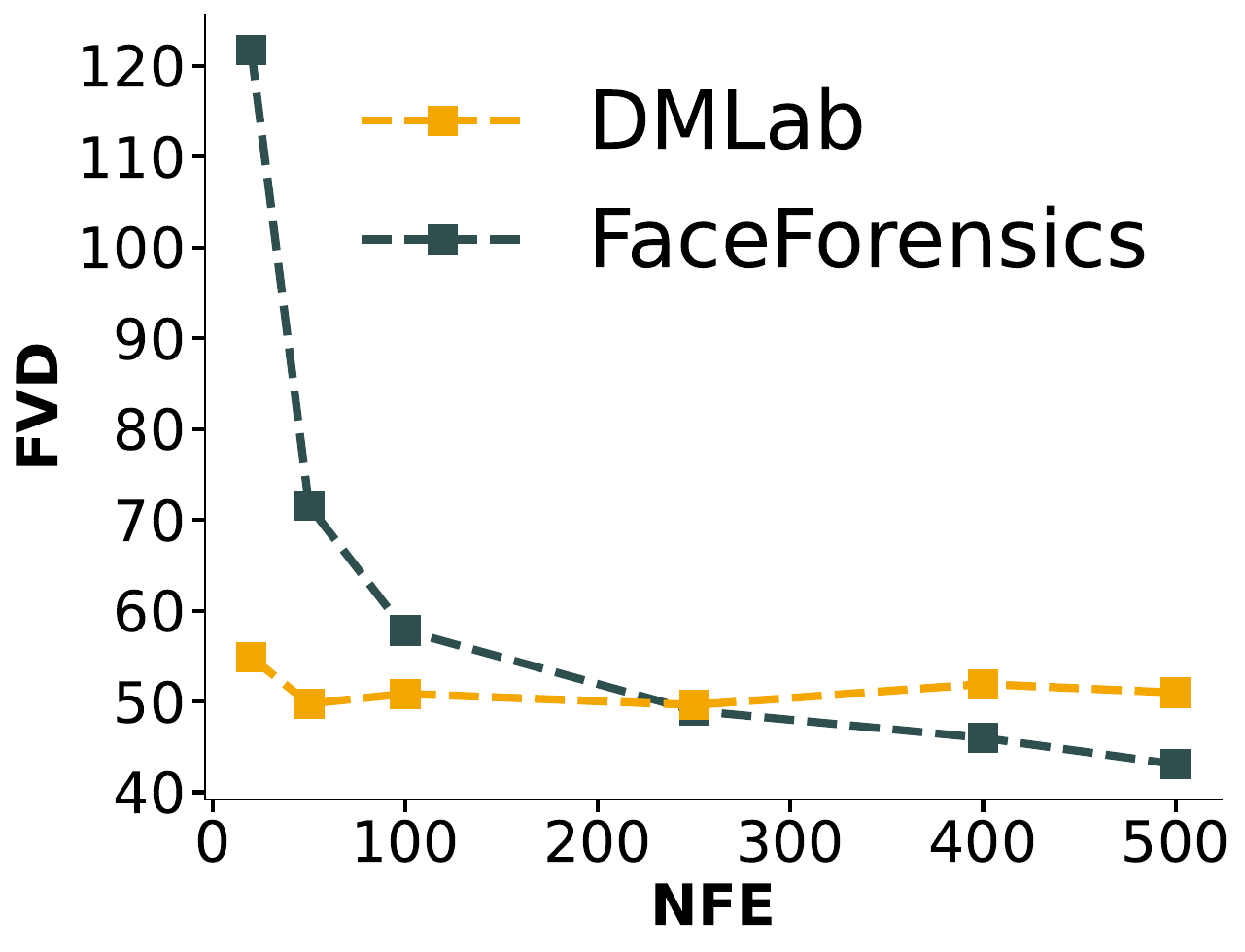}
        \caption{FM-style sampling}
        \label{fig:nfe_fvd_fm}
    \end{subfigure}
    %\vspace{-7pt}
    \caption{\textbf{NFE choices for both MGM-style ($20$) and FM-style ($250$) suitably trade off sampling speed with visual quality.} Figures show FVD on a single chunk of size $k$ for timestep-dependent frame-level masking models.}
    \vspace{-10pt}
    \label{fig:nfe_fvd}
\end{figure}

\paragraph{MGM-style and FM-style NFE choices minimize visual quality and sampling efficiency tradeoffs.}
The choice of NFE in our work is driven empirically. We compare generation quality when generating a single chunk $k$ on both datasets and tune our NFE accordingly for FM-style and MGM-style sampling modes. We are aware that our observations regarding sampling speeds depend on the choice of NFE, so we compare video quality for a lower number of sampling steps for both sampling modes on both datasets. In Figure ~\ref{fig:nfe_fvd}, we show that our choices of 20 for MGM-style and 250 for FM-style sampling achieve the best trade-off between sampling efficiency and quality, since video quality saturates for higher NFE in both modes across both datasets.

%\vspace{-10pt}
\section{Conclusion}

We have presented a discrete flow matching framework for flexible long video generation, leveraging frame-level masking during training to enable flexible, efficient sampling. Our experiments demonstrate that this approach can generate high-quality videos beyond 10$\times$ the training window length, while substantially reducing sampling cost through MGM-style unmasking. Notably, our models can seamlessly switch between timestep-dependent (flow matching) and timestep-independent (MGM) sampling modes without additional training, offering a unified solution that supports both full-sequence rollout and fully autoregressive generation. We believe discrete tokens have great potential for scalable visual generation. 

\section{Acknowledgements}

Special thanks go out to Timy Phan for proof-reading and
providing helpful comments.
This project has been supported by the German Federal Ministry for Economic Affairs and Climate Action within the project “NXT GEN AI METHODS – Generative Methoden für Perzeption, Prädiktion und Planung”, the bidt project KLIMA-MEMES, Bayer AG, and the German Research Foundation (DFG) project 421703927. The authors gratefully acknowledge the Gauss Center for Supercomputing for providing compute through the NIC on JUWELS at JSC and the HPC resources supplied by the Erlangen National High Performance Computing Center (NHR@FAU funded by DFG).

\newpage

{
    \small
    \bibliographystyle{ieeenat_fullname}
    \bibliography{main}
}


\clearpage
\setcounter{page}{1}
\maketitlesupplementary

\renewcommand{\thetable}{S\arabic{table}}
\renewcommand{\thefigure}{S\arabic{figure}}

\clearpage
\appendix

\onecolumn 

\section{Appendix}
\tableofcontents 
\clearpage

\begin{table*}[t]
\centering
\small
\renewcommand{\arraystretch}{1.2}
\setlength{\tabcolsep}{6pt}
\begin{tabular*}{\textwidth}{@{\extracolsep{\fill}} 
  >{\centering\arraybackslash}m{3cm}  % Sampling Mode
  >{\centering\arraybackslash}m{1.5cm} % Stride
  >{\centering\arraybackslash}m{3cm}   % Extrapolation Factor
  >{\centering\arraybackslash}m{2cm}   % Total NFE
  >{\centering\arraybackslash}m{3cm}   % Sampling Time [s]
  >{\centering\arraybackslash}m{1cm}   % FVD subcolumn 1
  >{\centering\arraybackslash}m{1cm}}  % FVD subcolumn 2
\toprule
\makecell{\textbf{Sampling}\\\textbf{Mode}} & 
\makecell{\textbf{Stride}} & 
\makecell{\textbf{Extrapolation}\\\textbf{Factor}} & 
\makecell{\textbf{Total}\\\textbf{NFE}} & 
\makecell{\textbf{Sampling}\\\textbf{Time [s]}} & 
\multicolumn{2}{c}{\textbf{FVD$\downarrow$}} \\
\cmidrule(rr){6-7}
 & & & & & \textbf{DMLab} & \textbf{FFS} \\
 \midrule
\rowcolor{gray!8}Diffusion Forcing~\cite{chen2024diffusionforcing} & $s=k-m$ & $1\times$ & $286 / 266$ & $45.32$ / $52.26$ & $60.30$ & $51.90$ \\
Rolling Diffusion~\cite{ruhe2024rollingdiffusionmodels} & $s=k-m$ & $1\times$ & $500$ / $500$ & $79.24$ / $98.23$ & \textbf{52.43} & \textbf{45.51} \\
\rowcolor{gray!8}\textit{MaskFlow} (MGM-Style) & $s=k-m$ & $1\times$ & \textbf{20} / \textbf{20} & \textbf{3.17 / 3.93} & $53.17$ & $45.92$ \\
\midrule
Diffusion Forcing~\cite{chen2024diffusionforcing} & $s=k-m$ & $2\times$ & $858$ / $798$ & $135.97$ / $156.78$ & $175.01$ & $144.43$ \\
\rowcolor{gray!8}Rolling Diffusion~\cite{ruhe2024rollingdiffusionmodels} & $s=k-m$ & $2\times$ & $896$ / $788$ & $141.99 / 154.81$ & 201.70 & 72.49 \\
\textit{MaskFlow} (MGM-Style) & $s=k-m$ & $2\times$ & \textbf{60} / \textbf{60} & \textbf{9.51} / \textbf{9.30} & $188.02$ &  $59.93$ \\
\rowcolor{gray!8}\textit{MaskFlow} (MGM-Style) & $s=1$ & $2\times$ & $740$ / $340$ & 117.27 / 66.80 & \textbf{50.87} & \textbf{30.43} \\
\midrule
Diffusion Forcing~\cite{chen2024diffusionforcing} & $s=k-m$ & $5\times$ & $2{,}002$ / $1{,}596$ & $317.27$ / $313.56$ & $232.89$ & $272.14$ \\
\rowcolor{gray!8}Rolling Diffusion~\cite{ruhe2024rollingdiffusionmodels} & $s=k-m$ & $5\times$ & $2{,}084$ / $1{,}652$ & $330.27$ / $324.56$ & $338.34$ & $248.13$ \\
\textit{MaskFlow} (MGM-Style) & $s=k-m$ & $5\times$ & \textbf{140} / \textbf{120} & \textbf{22.19 / 23.58} & $334.15$ & $108.74$ \\
\rowcolor{gray!8}\textit{MaskFlow} (MGM-Style) & $s=1$ & $5\times$ & $2{,}900$ / $1{,}300$ & 100.09/379.91 & \textbf{181.11} & \textbf{103.69} \\
\bottomrule
\end{tabular*}
\caption{\textbf{MGM Style sampling is much faster without sacrificing quality.} We report the total number of function evaluations (NFE), sampling time (in seconds), and FVD for various sampling methods and extrapolation factors across both datasets.}
\label{tab:speed_comparison}
\end{table*}

\subsection{Additional Related Work}

\paragraph{Masked Diffusion Models.} 
Limitations of autoregressive models for probabilistic language modeling have recently sparked increasing interest in masked diffusion models. Recent works like \cite{shi2024simplifiedgeneralizedmaskeddiffusion} and \cite{sahoo2024simpleeffectivemaskeddiffusion} have aligned masked generative models with the design space of diffusion models by formulating continuous-time forward and sampling processes. Works like \cite{nie2024scalingmaskeddiffusionmodels} and \cite{gong2024scalingdiffusionlanguagemodels} also demonstrate the significant scaling potential of MDM for language tasks, indicating that this masked modeling paradigm can rival autoregressive approaches for modalities beyond language such as protein co-design \cite{campbell2024generative} and vision.

\subsection{Computation of NFE for Different Sampling Methods}

Our sampling speed evaluations are determined by computing the required number of chunks 
\[
\ell = \left\lceil \frac{L - k}{s} \right\rceil + 1,
\]
to generate a video of total length \(L\), where \(k\) is the chunk size and \(s\) is the stride with which the chunk start is shifted. The overall number of function evaluations (NFEs) is then obtained by multiplying \(\ell\) with the number of sampling steps required to generate one chunk. We apply this methodology for all chunkwise-autoregressive approaches.

\begin{itemize}
    \item \textbf{MGM-Style Sampling:} In this method each chunk is generated in $20$ forward passes, so that the total NFE is
    \[
    \text{NFE}_{\mathrm{MGM}} = \ell \times 20.
    \]

    \item \textbf{FM-Style Sampling:} Here we generate each chunk in $250$ forward passes:
    \[
    \text{NFE}_{\mathrm{FM}} = \ell \times 250.
    \]
    
    \item \textbf{Diffusion Forcing with Pyramid Scheduling:} Here, we apply $250$ sampling timesteps per frame but begin unmasking earlier frames as the denoising process proceeds. For a chunk of \(k\) frames, we generate a scheduling matrix with 
    \[
    H = 250 + (k-1) + 1 = k + 250
    \]
    rows and \(k\) columns. Each entry in the scheduling matrix is computed as
    \[
    \text{scheduling\_matrix}[i,j] = 250 + j - i,\quad \text{for } i=0,\ldots,H-1 \text{ and } j=0,\ldots,k-1,
    \]
    and then clipped to the interval \([0,249]\). Since we iterate through each of the $H$ rows of the denoising matrix in each chunk we effectively compute
    \[
    \text{NFE}_{\text{DiffusionForcing}} = k + 250.
    \] 
    
    \item \textbf{RDM Sampling:} This approach proceeds in three stages:
    \begin{enumerate}
        \item \textit{Initialization (Init-Schedule):} The initial window of \(k\) frames is processed using a fixed schedule that applies $T=250$ forward passes to bring the window to its rolling state.
        
        \item \textit{Sliding Window Handling:} After initialization, the window is shifted by one frame at a time. For each shift, an inner loop is executed that updates the denoising levels until the first non-context frame (i.e., the frame immediately following the \(m\) context frames) is fully denoised (i.e., reaches a value of 1). This inner loop requires $\left\lceil \frac{T}{k-m} \right\rceil$ forward passes per window shift. As the window is shifted \((L - k)\) times, this stage contributes roughly \((L - k) \times \left\lceil \frac{T}{k-m} \right\rceil\) forward passes.
        
        \item \textit{Final Window Processing:} Once the sliding window stage is complete, the final (partial) window is further refined until all frames are fully denoised. This final stage requires additional $250$ forward passes.
    \end{enumerate}
    
    Thus, the total NFE for RDM is given by
    \[
    \text{NFE}_{\mathrm{Rolling}} = 250 \; (\text{init-schedule}) + (L - k) \times \left\lceil \frac{T}{k-m} \right\rceil\ \; (\text{sliding}) + 250 \; (\text{final window}).
    \]
\end{itemize}

\subsection{Training \& Implementation Details}

All FFS models were trained on 4 H100 GPUs with a local batch size of $4$. We run training for a total of $200{,}000$ steps and use a sigmoid scheduler that determines the per-frame masking ratio for a sampled masking level $t^k$. We use an AdamW optimizer with a learning rate of $1e-4$ and $\beta_1 = 0.9$ and $\beta_2 = 0.999$. We additionally incorporate a frame-level loss weighting mechanism based that is also based on \(t^k\). We adopt \emph{fused}-SNR loss weighting from \cite{hang2023efficient,chen2024diffusionforcing} and derive it for discrete flow matching. Let

\[
\text{SNR}(t) \;=\; \frac{\kappa(t)^2}{\,1 - \kappa(t)^2\,},
\]

where \(\kappa(t)\) is the masking schedule. The \emph{fused}-SNR mechanism smoothes SNR values across time steps in a video by computing an exponentially decaying SNR from previous frames (or tokens). We refer the reader to~\cite{chen2024diffusionforcing} for full details.

\begin{algorithm}[!ht]
\caption{\textbf{FM-Style Sampling with Context Frames for a Single Chunk}}
\label{alg:fmsampling}
\begin{algorithmic}[1]
\REQUIRE 
   $p(\mathbf{x}_1 | \mathbf{x}_t, \mathbf{t};\theta)$, 
   $t$, 
   context frames $\mathbf{c} = (c^1,\dots,c^m)$, 
   fully masked frame \([M]\) (i.e., a frame where every token equals the mask token \(M\)),
   $t \in [0,1]$, 
   $\Delta t$

\STATE $\mathbf{x}_t \,\gets\, (\,c^1,\dots,c^m,\,[M],\dots,[M])$
\STATE $t \,\gets\, 0$
\STATE $\mathbf{t} \gets (1,\dots,1,0,\dots0)$

\WHILE{$t \,\le\, 1 - \Delta t$}
    \STATE $u_t(\mathbf{x}_t) 
        \;=\; 
        \frac{t}{1-t}
        \Bigl[
          p_\theta(\mathbf{x}_1 \mid \mathbf{x}_t,\,\mathbf{t}) 
          \;-\; 
          \delta_{\mathbf{x}_t}
        \Bigr]$
    \STATE $p_\theta\!\bigl(\mathbf{x}_1 \mid \mathbf{x}_{t+\Delta t},\,\mathbf{t}+\Delta t\bigr)
        \;=\;
        \mathrm{Cat}\!\Bigl[\,
          \delta_{\mathbf{x}_t}
          \;+\; 
          u_t(\mathbf{x}_t)\,\Delta t
        \Bigr]$
    \STATE \textbf{For each token} $n$ in $\mathbf{x}_t$: 
    \STATE \quad 
    $
       x_{t+\Delta t}^{n} \gets
       \begin{cases}
          x_t^{n}, & \text{if } x_t^{n} \neq M,\\
          p(\cdot | \mathbf{x}_{t+\Delta t},\,\mathbf{t}+\Delta t; \theta), & \text{if } x_t^{n} = M.
       \end{cases}
    $
\STATE $t \gets t + \Delta t$
\STATE $\mathbf{t} \gets \mathbf{t} + \Delta t$

\ENDWHILE
\STATE \textbf{return} $\mathbf{x}_t$
\end{algorithmic}
\end{algorithm}

\begin{algorithm}[ht]
\caption{\textbf{MGM-Style Sampling for a Single Chunk}}
\label{alg:mgm_chunk_unmasking_revised}
\begin{algorithmic}[1]
\REQUIRE 
  Network $p(\mathbf{x}_1 \mid \mathbf{x}_t, \mathbf{t}; \theta)$,  
  context frames $\mathbf{c} = (c^1,\dots,c^m)$,  
  masked frame $[M]$ (i.e., every token equals $M$),    
  total unmasking steps $T$
\STATE \textbf{Initialize:}\\
$\mathbf{x}_t \;\leftarrow\; (\mathbf{c},\, [M],\dots,[M])$\\
$\mathbf{t} \;\leftarrow\; (\underbrace{1,\dots,1}_{m},\, \underbrace{0,\dots,0}_{k-m})$
\STATE Define the set of masked token indices in $\mathbf{x}_t$:\\
$\mathcal{M} \;\triangleq\; \{\, n \mid x_t^n = M \,\}.$
\FOR{$i=1$ \textbf{to} $T$}
    \STATE Compute token-wise logits:\\
    $\boldsymbol{\lambda} \;\leftarrow\; p(\mathbf{x}_1 \mid \mathbf{x}_t, \mathbf{t}; \theta).$
    \STATE \textbf{For each token} $n \in \mathcal{M}$: \\
    sample $\hat{x}_t^n \sim \mathrm{Cat}\Bigl(\mathrm{Softmax}\bigl(\boldsymbol{\lambda}^n\bigr)\Bigr)$ \\
    and compute the confidence score 
    $C_n \;=\; \mathrm{Softmax}\bigl(\boldsymbol{\lambda}^n\bigr)_{\hat{x}_t^n}.$ \\
    \STATE \textbf{Define the confidence threshold:}\\
    Let $\alpha$ denote the desired fraction of masked tokens to update in each iteration (e.g. $\alpha = 1/T$). \\
    
    Then set 
    $\tau_c \;=\; \min\Bigl\{ c \in [0,1] \;\Bigm|\; \Bigl|\{ j \in \mathcal{M} \mid C_j \ge c \}\Bigr| \ge \Bigl\lceil \alpha\,|\mathcal{M}| \Bigr\rceil \Bigr\}.$ \\
    
    (That is, $\tau_c$ is chosen as the minimum confidence such that at least $\lceil \alpha\,|\mathcal{M}| \rceil$ tokens have confidence scores at or above $\tau_c$, thereby selecting the top $\lceil \alpha\,|\mathcal{M}| \rceil$ tokens.)
    \STATE \textbf{For each token} $n \in \mathcal{M}$ with $C_n \ge \tau_c$, update:\\
    $x_t^n \;\leftarrow\; \hat{x}_t^n.$
    \STATE Update the set of masked indices:\\
    $\mathcal{M} \;\leftarrow\; \{\, n \mid x_t^n = M \,\}.$
    \IF{$\mathcal{M} = \varnothing$}
         \STATE \textbf{break}
    \ENDIF
\ENDFOR
\STATE \textbf{return} $\mathbf{x}_t$.
\end{algorithmic}
\end{algorithm}

\subsection{Baseline Details}

The two most comparable works to our method are \citet{chen2024diffusionforcing} and \citet{ruhe2024rollingdiffusionmodels}. Both of these techniques propose novel sampling methods that can be rolled out to long video lengths, and also apply frame-specific noise levels. Both of these approaches are diffusion-based and operate on continuous representations, whereas we operate on discrete tokens and use masking. We re-implement both the pyramid sampling scheme proposed in Diffusion Forcing and the Rolling Diffusion sampling method in our discrete setting. This allows us to compare the baseline sampling methods to MaskFlow on the same model backbones. 
To isolate the effect of our chunkwise autoregressive sampling methodology on performance from the effects of tokenization, we reimplement both the pyramid sampling scheme proposed in Diffusion Forcing and the Rolling Diffusion sampling method for our discrete setting. This allows us to compare the baseline sampling methods on the same timestep-dependent model backbone. 
Although it is conceivable that Rolling Diffusion sampling may perform better when applied to a model explicitly trained using the progressive noise schedule suggested in \citet{ruhe2024rollingdiffusionmodels}, we believe this comparison is still fair. Our training methodology does not inject any inductive bias by way of the masking level into the model, so there is no obvious advantage that our sampling should have over other methods. 
We provide a comprehensive evaluation of performance and sampling efficiency across both datasets and different sampling modes.

\subsection{Dataset Details}

\paragraph{Deepmind Lab.} The Deepmind Lab (DMLab) navigation dataset contains $64 \times 64$ resolution videos of random walks in a 3D maze environment. We use the total 625 videos with frame length 300 frames, and randomly sample sequences of 36 consecutive frames from each video during training. We upscale video frames to a resolution of $256 \times 256$ before tokenizing them similar to our approach for FaceForensics. We disregard the provided actions, focusing on action-unconditional video generation. We use $m=12$ and $s=24$ for the DMLab full sequence generation experiments unless stated otherwise.

\paragraph{FaceForensics.} FaceForensics (FFS) is a dataset that contains $150\times150$ images of deepfake faces, totaling 704 videos with varying number of frames at 8 frames-per-second. We upsample the resolution to $256 \times 256$, before encoding individual frames using the image-based tokenizer SD-VQGAN \cite{rombach2022high_latentdiffusion_ldm}. While image-based tokenizers have shown to lead to flickering issues, we observe high-reconstruction quality (reconstruction FVD $\approx 8$ on FFS) on our datasets and thus leave work on video tokenization to other works. After tokenization, we train on encoded frame sequences of 16 frames, each consisting of token grids with dimensionality $32 \times 32$. We generally use $m=2$ ground-truth context frames for conditioning, and $s=14$.

\subsection{Further Quantitative Results}

\paragraph{Our chunkwise autoregressive MGM-style sampling is preferable to full sequence training in settings with limited hardware.} To evaluate our method for long video generation against a longer training window baseline, we compare the performance of a frame-level masking model trained on $16$ frames with full sequence generation of a constant-masking level model trained on $32$ frames with similar batch size and on similar hardware. In Table ~\ref{tab:longer_train_window_baseline} we show that iterative rollout of our MGM-style sampling outperforms full sequence generation even when the full sequence model is trained on a longer window.

\begin{table}[ht]
    \centering
    \normalsize
    \resizebox{0.48\textwidth}{!}{%
    \begin{tabular}{l|cccc}
    \toprule
    \makecell{\textbf{Sampling} \\ \textbf{Mode}} 
    & \makecell{\textbf{Training} \\ \textbf{Window}} 
    & \makecell{\textbf{Sampling} \\ \textbf{Window}} 
    & \makecell{\textbf{Total} \\ \textbf{NFE}}
    & \makecell{\textbf{FVD} $\downarrow$} \\
    \midrule
    FM-Style (bs=2) & 32 & 32 & 250 & 253.08 \\
    \midrule
    \textit{MaskFlow} (MGM-Style) (bs=2) & 16 & 32 & 60 & 192.76 \\
    \rowcolor{gray!8}\textit{MaskFlow} (MGM-Style) (bs=4) & 16 & 32 & 60 & \textbf{59.93} \\
    \bottomrule
    \end{tabular}
    }
    \caption{\textbf{Our MGM-style sampling is more efficient and generates better results over baseline for larger training windows}. We train a constant masking ratio model on larger window sizes with similar batch size on similar hardware, and compare full sequence generation to generating the same length using our chunkwise MGM-style sampling.}
    \label{tab:longer_train_window_baseline}
\end{table}

\begin{table}[ht]
    \centering
    \normalsize
    \resizebox{0.48\textwidth}{!}{%
    \begin{tabular}{l|ccrr}
        \toprule
        & \makecell{\textbf{Extrapolation} \\ \textbf{Factor}}
        & \makecell{\textbf{Sampling} \\ \textbf{Stride}}
        & \makecell{\textbf{Total} \\ \textbf{NFE}}
        & \makecell{\textbf{FVD} $\downarrow$} \\
        \midrule
        FaceForensics   & $2\times$  & $s=14$ (\textit{full sequence}) & \textbf{60} & 59.93 \\
        \rowcolor{gray!8}FaceForensics   & $2\times$  & $s=1$ (\textit{autoregressive})  & 340 & \textbf{30.43} \\
        \midrule
        FaceForensics   & $5\times$  & $s=14$ (\textit{full sequence}) & \textbf{120} & 108.74 \\
        \rowcolor{gray!8}FaceForensics & $5\times$  & $s=1$ (\textit{autoregressive})  & 1,300 & \textbf{103.69} \\
        \midrule
        FaceForensics   & $10\times$ & $s=14$ (\textit{full sequence}) & \textbf{240} & 214.39 \\
       \rowcolor{gray!8} FaceForensics   & $10\times$ & $s=1$ (\textit{autoregressive})  & 2,900 & \textbf{165.02} \\
        \midrule
        \midrule
        DMLab & $2\times$  & $s=24$ (\textit{full sequence}) & \textbf{60} & 188.22 \\
        \rowcolor{gray!8}DMLab & $2\times$  & $s=1$ (\textit{autoregressive})  & 740  & \textbf{50.87} \\
        \midrule
        DMLab & $5\times$  & $s=24$ (\textit{full sequence}) & \textbf{140}  & 334.15 \\
       \rowcolor{gray!8} DMLab & $5\times$  & $s=1$ (\textit{autoregressive})  & 2,900 & \textbf{181.11} \\
        \bottomrule
    \end{tabular}
    }
    \caption{\textbf{Autoregressive sampling outperforms full sequence sampling on timestep-dependent models at the cost of higher NFE.}}
    \label{tab:autoregression_dependent}
\end{table}

\begin{figure*}[ht!]
    \centering
    \includegraphics[width=0.7\textwidth]{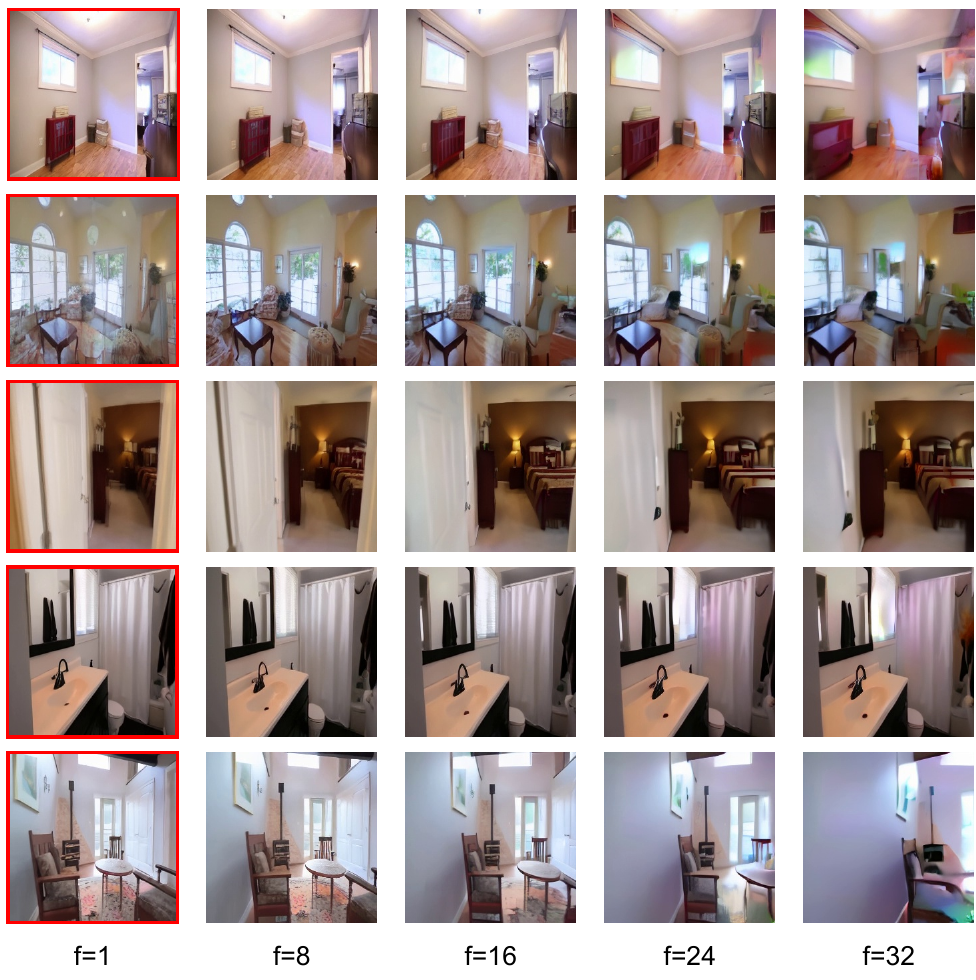}\hfill
    \caption{\textbf{Further visualizations on the Realestate10K \cite{zhou2018stereo} dataset.} Models trained on chunk size $k = 16$ with $4$ H100 GPUs. Due to computational limitations, we cannot  provide further analyses on this larger, more compute intensive dataset.}
    \label{fig:faces_comparison}
\end{figure*}

\begin{table}[ht]
    \centering
    \normalsize
    \resizebox{0.58\textwidth}{!}{%
    \begin{tabular}{l|ccrr}
        \toprule
        & \makecell{\textbf{Extrapolation} \\ \textbf{Factor}}
        & \makecell{\textbf{Sampling} \\ \textbf{Stride}}
        & \makecell{\textbf{Total} \\ \textbf{NFE}}
        & \makecell{\textbf{FVD} $\downarrow$} \\
        \midrule
        FaceForensics   & $2\times$  & $s=14$ (\textit{full sequence}) & \textbf{60} & 109.96 \\
        FaceForensics   & $2\times$  & $s=1$ (\textit{autoregressive}) & 340 & \textbf{43.91} \\
        \midrule
        FaceForensics   & $5\times$  & $s=14$ (\textit{full sequence}) & \textbf{120} & \textbf{137.66} \\
        FaceForensics & $5\times$  & $s=1$ (\textit{autoregressive})  & 1,300 & 193.90 \\
        \midrule
        FaceForensics   & $10\times$ & $s=14$ (\textit{full sequence}) & \textbf{240} & \textbf{174.92} \\
        FaceForensics   & $10\times$ & $s=1$ (\textit{autoregressive})  & 2,900 & 293.16 \\
        \midrule
        \midrule
        DMLab & $2\times$  & $s=24$ (\textit{full sequence}) & \textbf{60} & 219.33 \\
        DMLab & $2\times$  & $s=1$ (\textit{autoregressive})  & 740  & \textbf{42.53} \\
        \midrule
        DMLab & $5\times$  & $s=24$ (\textit{full sequence}) & \textbf{140}  & 402.73 \\
        DMLab & $5\times$  & $s=1$ (\textit{autoregressive})  & 2,900 & \textbf{80.56} \\
        \bottomrule
    \end{tabular}
    }
    \caption{\textbf{Autoregressive sampling outperforms full sequence sampling on timestep-independent models at the cost of higher NFE.} Performance improvement on DMLab is substantial.}
    \label{tab:autoregression_independent}
\end{table}

\newpage

\subsection{Further Qualitative Results}

\begin{figure*}[ht!]
    \centering
    \includegraphics[width=0.48\textwidth]{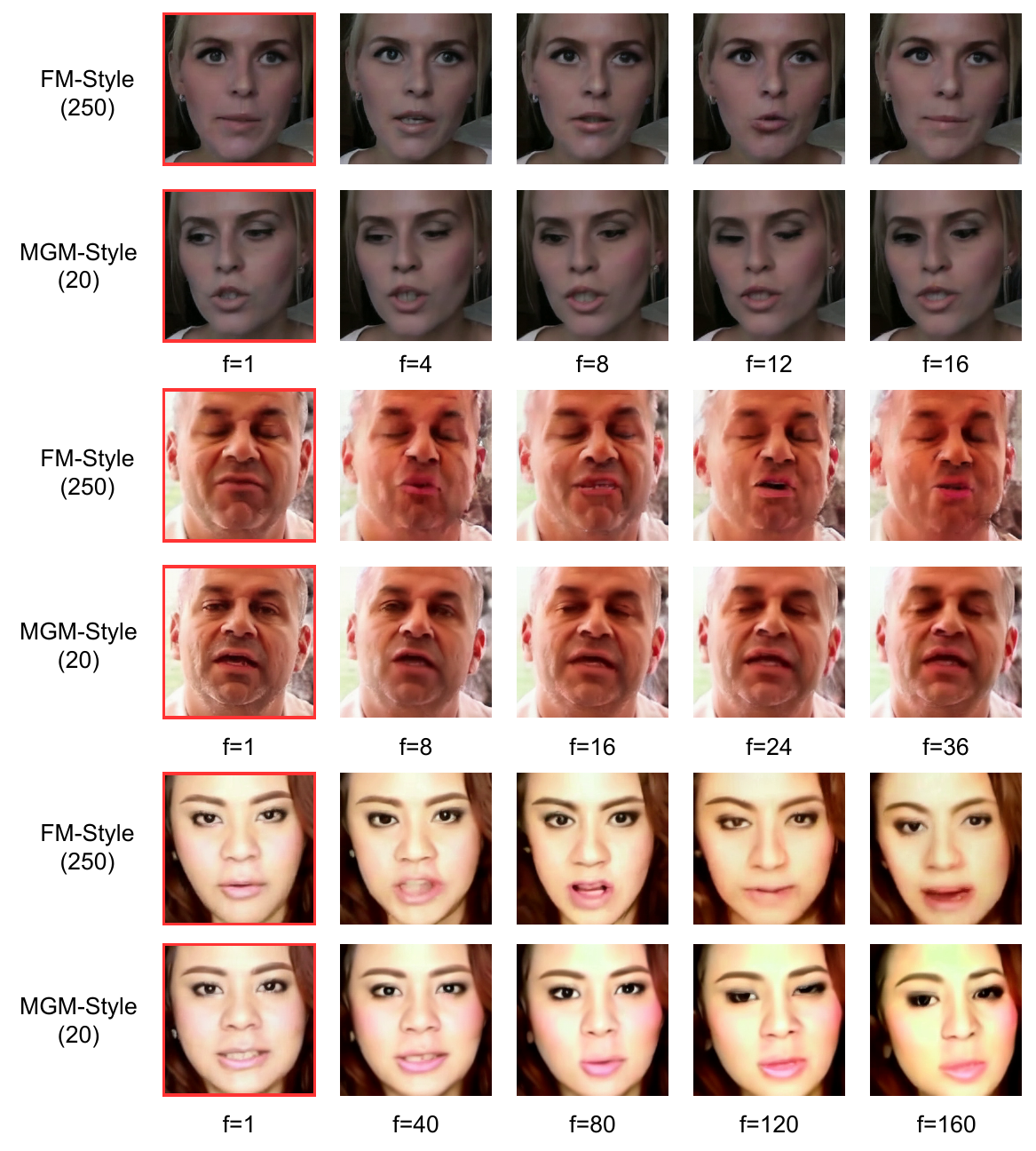}\hfill
    \includegraphics[width=0.48\textwidth]{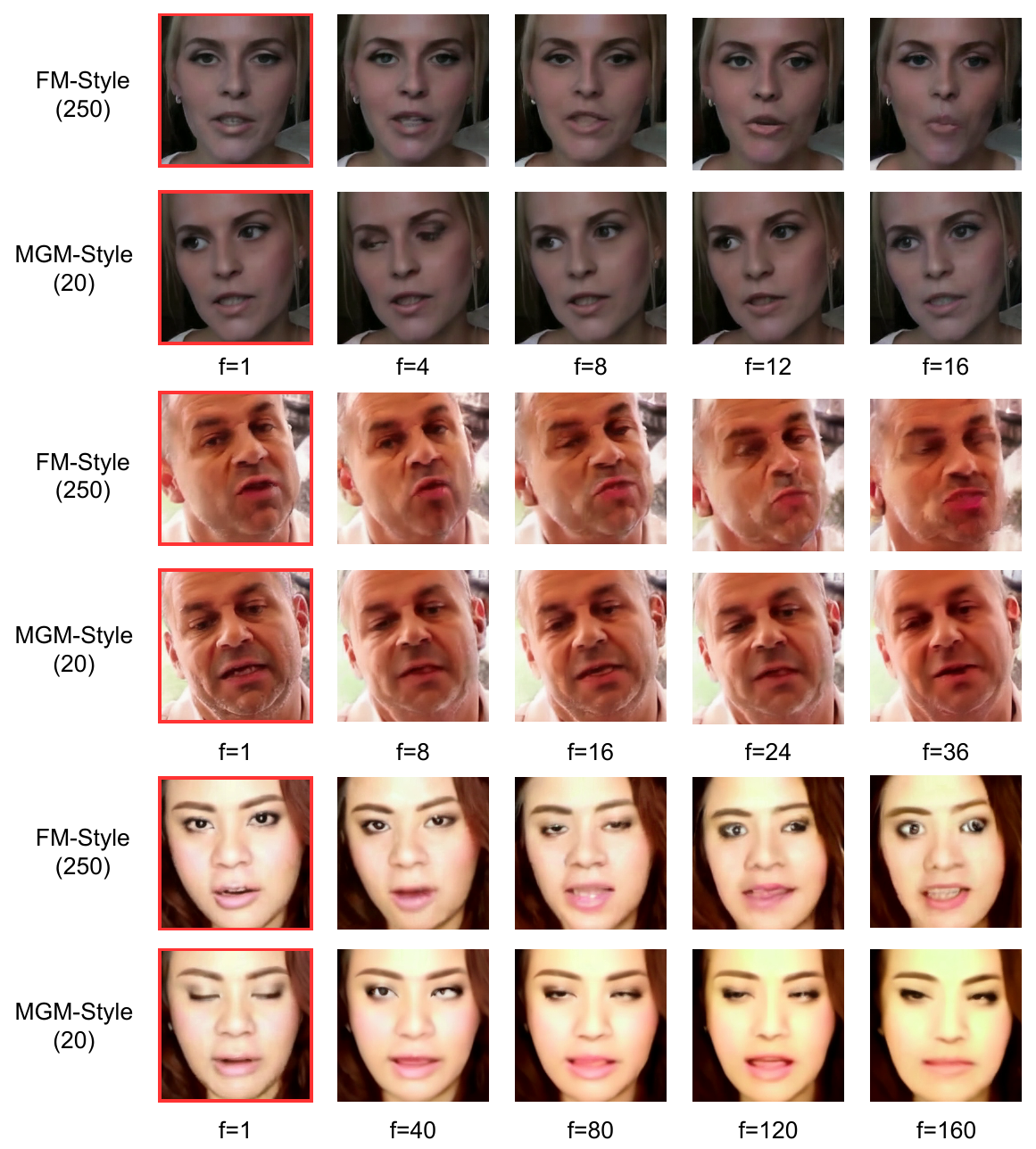}
    \caption{\textbf{Visualizations of FaceForensics generation results with different context frames.}}
    \label{fig:faces_comparison}
\end{figure*}

\end{document}